\documentclass[sigconf]{acmart}

\usepackage{dsfont}
\usepackage{multirow}
\usepackage{bm}
\newcommand{\figref}[1]{Figure \ref{#1}}
\newcommand{\tabref}[1]{Table \ref{#1}}
\newcommand{\secref}[1]{Section \ref{#1}}

\newcommand{\equref}[1]{Equation (\ref{#1})}
\usepackage{multirow}
\usepackage[normalem]{ulem}
\useunder{\uline}{\ul}{}

\AtBeginDocument{%
  \providecommand\BibTeX{{%
    \normalfont B\kern-0.5em{\scshape i\kern-0.25em b}\kern-0.8em\TeX}}}

\acmDOI{10.1145/3637528.3672001}

\copyrightyear{2024}
\acmYear{2024}
\setcopyright{rightsretained}
\acmConference[KDD '24] {Proceedings of the 30th ACM SIGKDD Conference on Knowledge Discovery and Data Mining }{August 25--29, 2024}{Barcelona, Spain.}
\acmBooktitle{Proceedings of the 30th ACM SIGKDD Conference on Knowledge Discovery and Data Mining (KDD '24), August 25--29, 2024, Barcelona, Spain}
\acmISBN{979-8-4007-0490-1/24/08}

\begin{document}

\title{Repeat-Aware Neighbor Sampling for Dynamic Graph Learning}

\author{Tao Zou}
\email{zoutao@buaa.edu.cn}
\affiliation{
  \institution{CCSE Lab, Beihang University}
  \city{Beijing}
  \country{China}
  \postcode{100191}
}

\author{Yuhao Mao}
\email{maoyuhao@buaa.edn.cn}
\affiliation{
  \institution{CCSE Lab, Beihang University}
  \city{Beijing}
  \country{China}
  \postcode{100191}
}

\author{Junchen Ye}
\authornote{Corresponding Author}
\email{junchenye@buaa.edu.cn}
\affiliation{
  \institution{School of Transportation Science and Engineering, Beihang University}
  \city{Beijing}
  \country{China}
}

\author{Bowen Du}
\email{dubowen@buaa.edu.cn}

\affiliation{
  \institution{School of Transportation Science and Engineering, Beihang University}
  \institution{Zhongguancun Laboratory}
   \institution{CCSE Lab}
  \city{Beijing}
  \country{China}
}


\renewcommand{\shortauthors}{Tao Zou, Yuhao Mao, Junchen Ye, and Bowen Du.}



\begin{abstract}
Dynamic graph learning equips the edges with time attributes and allows multiple links between two nodes, which
is a crucial technology for understanding evolving data scenarios like traffic prediction and recommendation systems.
Existing works obtain the evolving patterns mainly depending on the most recent neighbor sequences. 
However, we argue that whether two nodes will have interaction with each other in the future is highly correlated with the same interaction that happened in the past. Only considering the recent neighbors overlooks the phenomenon of repeat behavior and fails to accurately capture the temporal evolution of interactions.
To fill this gap, this paper presents RepeatMixer, which considers evolving patterns of first and high-order repeat behavior in the neighbor sampling strategy and temporal information learning. 
Firstly, we define the first-order repeat-aware nodes of the source node as the destination nodes that have interacted historically and extend this concept to high orders as nodes in the destination node's high-order neighbors. Then, we extract neighbors of the source node that interacted before the appearance of repeat-aware nodes with a slide window strategy as its neighbor sequence. Next, we leverage both the first and high-order neighbor sequences of source and destination nodes to learn temporal patterns of interactions via an MLP-based encoder. 
Furthermore, considering the varying temporal patterns on different orders, we introduce a time-aware aggregation mechanism that adaptively aggregates the temporal representations from different orders based on the significance of their interaction time sequences. Experimental results demonstrate the superiority of RepeatMixer over state-of-the-art models in link prediction tasks, underscoring the effectiveness of the proposed repeat-aware neighbor sampling strategy.
\end{abstract}



\maketitle

\section{Introductioin}
\label{section-1}
Dynamic graph learning has been employed in various scenarios with evolving graph input data, such as recommendation systems \cite{DBLP:conf/kdd/ZhengWWL022}, patent applicant trend prediction \cite{DBLP:journals/tkde/ZouYSDWZ24} and social networks \cite{DBLP:conf/cikm/SharmaRLMK23, DBLP:conf/kdd/KumarZL19}. To capture the fine-grained temporal information, existing works \cite{DBLP:conf/kdd/KumarZL19, DBLP:journals/corr/abs-2006-10637, DBLP:conf/nips/0004S0L23} treat dynamic graphs as sequences of timestamped interactions arranged in chronological order and derive node representations from their historical neighbor sequences.

\begin{figure}[!htbp]
    \centering
\includegraphics[width=1.00\columnwidth]{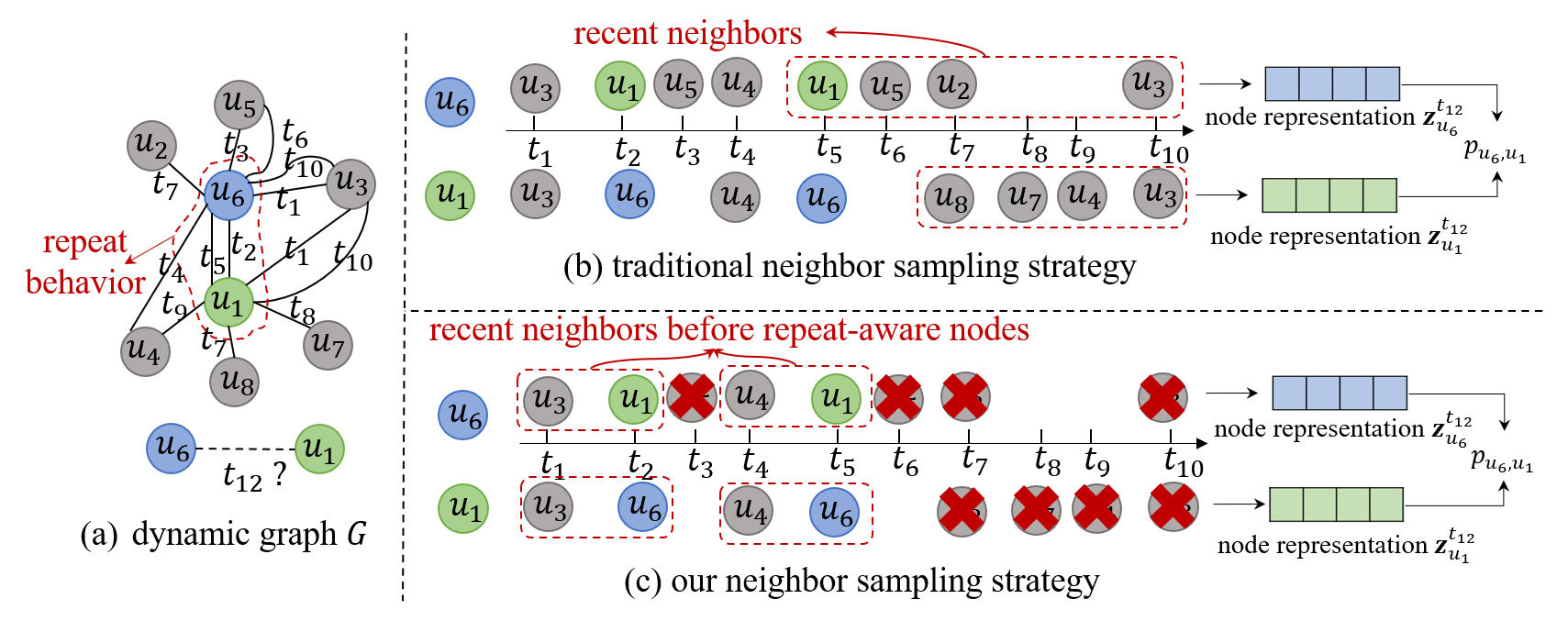}
    \caption{We show a dynamic graph $G$ evolves from $t_0$ to $t_6$ in (a). Notably, some interactions occur multiple times, such as the interaction between $u_6$ and $u_1$. Now we aim to predict whether $u_6$ will interact with $u_1$ at timestamp $t_9$. To generate the temporal representations of $u_6$ and $u_1$, we obtain neighbor sequences via sampling strategies (b) and (c).}
    \label{fig:dynamic graph}
\end{figure}

Recently, various approaches have been proposed in dynamic graph learning based on memory-based networks \cite{DBLP:conf/kdd/KumarZL19, DBLP:journals/corr/abs-2006-10637}, temporal random walks \cite{DBLP:conf/kdd/YuCAZCW18}, and sequential models \cite{DBLP:journals/corr/abs-2105-07944, DBLP:conf/nips/0004S0L23}. Initially, they extract the most recent neighbors as neighbor sequences and learn node representations by either modeling temporal patterns from each neighbor sequence or integrating correlations between two nodes' sequences. However, we argue that the node representations are highly correlated with the temporal patterns of the same interactions (i.e., repeat behavior \cite{DBLP:conf/recsys/Reiter-HaasPSMT21,DBLP:conf/sigir/AriannezhadJLFS22}). For example, we aim to predict whether $u_6$ will interact with $u_1$ at timestamp $t_9$ in \figref{fig:dynamic graph} (a). Existing works tend to sample the most recent neighbors of $u_6$ and $u_1$ for capturing temporal patterns, which causes $u_1$ to overlook the changing details of the same interaction that occurred at $t_2$, as depicted in \figref{fig:dynamic graph} (b). Hence, \textit{how to design an effective neighbor sampling strategy (NSS) considering the temporal evolution of interactions is an important issue in dynamic graph learning}.


 \begin{figure}[!htbp]
    \centering
\includegraphics[width=1.00\columnwidth]{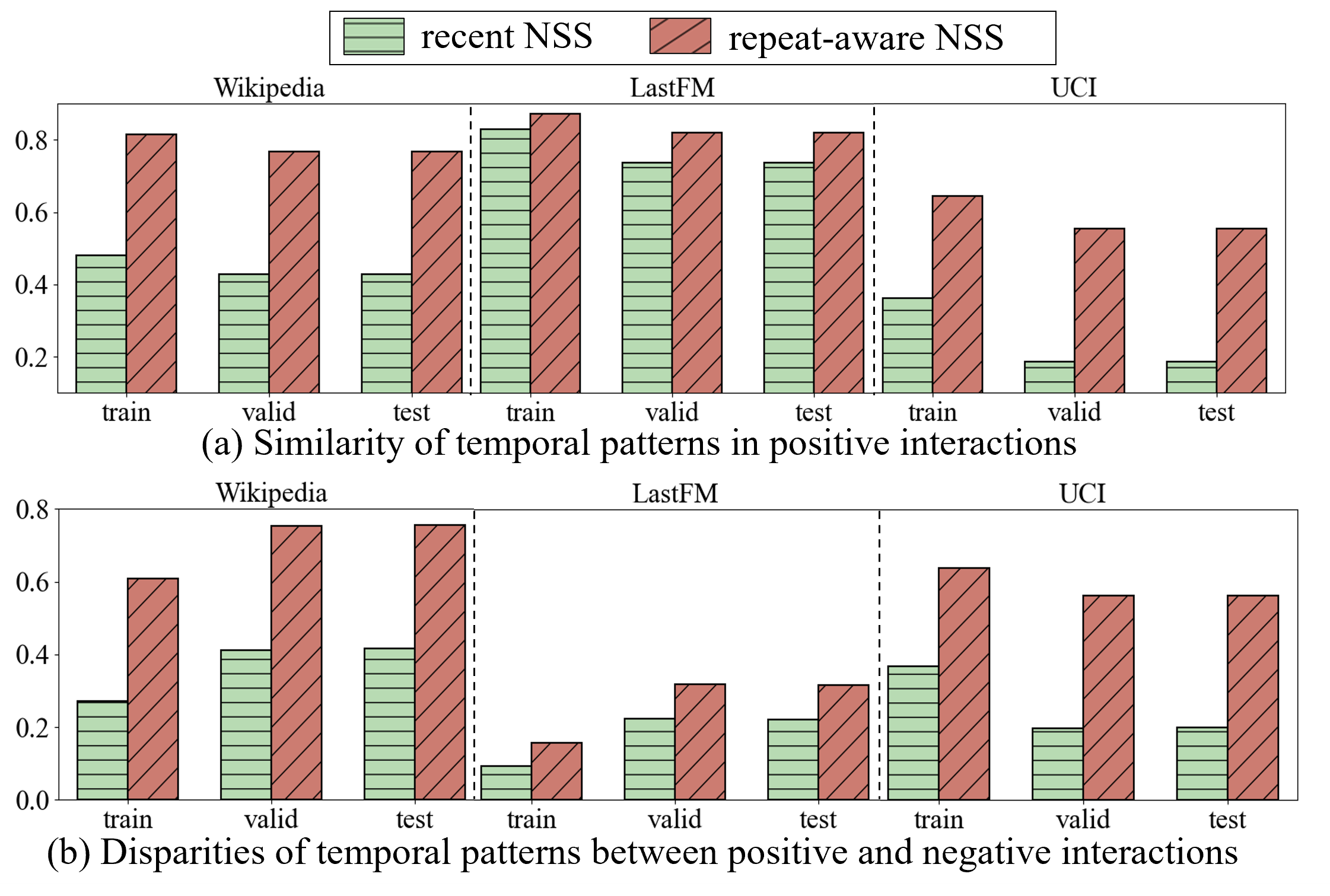}
    \caption{We show the average similarity of positive interactions in (a) and the average discrepancy between positive and negative interactions in (b) on three datasets. Each positive interaction is associated with a random negative interaction.}
    \label{fig:introduction}
\end{figure}

To evaluate the effect on repeat behavior, we conduct a pre-experiment and find that sampling neighbors associated with repeated behaviors assists models in acquiring more accurate temporal patterns between nodes. The settings of our experiment are as follows: given an interaction, we compare the similarity of temporal patterns between the source and target node by computing the pearson correlation coefficient (PCC) on their time interval sequences obtained from neighbor sequences. Specifically, for $(u, v, t)$, we sample $k$ recent neighbors for $u$ and $v$ until timestamp $t$ (i.e., recent NSS) or the recent neighbors appeared before the repeat interaction $e^\prime=(u,v,t_k)$ (where $t_k<t$) (i.e., repeat-aware NSS) to form neighbor sequences. Then we compute the time interval sequences by determining the time intervals of interactions until the current timestamp $t$. Finally, we get the similarity scores for interactions in \figref{fig:introduction}. Based on the results, we note that the temporal correlation among positive interactions and the discrepancies between positive and negative interactions are more pronounced in comparison to the recent NSS. This suggests that considering the evolution of current interactions in the neighbor sampling process helps us learn correlated temporal patterns between nodes.


To tackle the problem, we propose RepeatMixer, a dynamic graph learning method that considers the temporal patterns of first and high-order repeat behavior in the neighbor sampling strategy and learning temporal representations for interactions. Firstly, we design a repeat-aware neighbor sampling strategy, which selects the most recent neighbors that appeared before repeat-aware nodes with a slide window strategy as its neighbor sequences. Specifically, we define the first-order repeat-aware nodes of the source node as the destination nodes that appeared in the past and we extend the concept into high orders to capture high-order temporal information. Then we employ an MLP-based encoder to capture the long-term correlated temporal information between two nodes based on their first and high-order neighbor sequences. Besides, to fuse the temporal patterns from different orders, we utilize a time-aware adaptive aggregation mechanism to aggregate their representations according to the significance of their interaction time sequences.
Experimental results show that our approach could consistently outperform the state-of-the-art on the link prediction task. We also provide an in-depth analysis of the repeat-aware neighbor sampling strategy and time-aware aggregation mechanism. We summarize the main contributions of this paper as follows.

\begin{itemize}

    \item A novel repeat-aware neighbor sampling strategy is proposed for dynamic graph learning. Unlike existing works that focus on capturing the node-wise temporal behaviors by recent interaction sequence, we consider the pair-wise temporal patterns and unleash the power of the repeat behavior by utilizing the neighbors of the same interaction that happened in the past.


    \item Correspondingly, the explicit definition of the repeat-aware neighbor is provided, along with a detailed discussion of the first and second-order repeat patterns. A novel aggregation mechanism is proposed to calculate the significant score and fuse information from different orders adaptively.

    
    \item Experiments on real-world datasets confirm the effectiveness of the proposed method. The enhanced performance on baselines equipped with our sampling strategy further demonstrates our versatility.
\end{itemize}

\section{Preliminaries}
\label{section-2}

\textbf{Definition 1. Dynamic Graph.} A dynamic graph is represented as a sequence of non-decreasing chronological interactions denoted by $\mathcal{G} = \{(u_1, v_1, t_1), (u_2, v_2, t_2), \ldots, (u_k, v_k, t_k)\}$, where $0 \leq t_1 \leq t_2 \leq \ldots \leq t_k$. In this representation, $u_i$ and $v_i$ signify the source node and destination node, respectively for the $i$-th interaction occurring at timestamp $t_i$. The set of all nodes is denoted by $\mathcal{N}$. Each node $u \in \mathcal{N}$ is associated with a node feature $\bm{x}_u \in \mathbb{R}^{d_N}$, and each interaction $(u,v,t)$ is characterized by an edge feature $\bm{x}^{t}_{u,v} \in \mathbb{R}^{d_E}$. Here, $d_N$ and $d_E$ represent the dimensions of the node feature and edge feature. For non-attributed graphs, we set the node feature and edge feature to zero vectors, i.e., $\bm{x}_u = \bm{x}_e = \mathbf{0}$.


\textbf{Definition 2. Problem Formalization.} Given the source node $u$, destination node $v$, timestamp $t$, and historical interactions before $t$, i.e., $\{(u^{\prime}, v^{\prime}, t^{\prime})|t^{\prime}<t\}$, representation learning on dynamic graph aims to design a model to learn time-aware representations $\bm{z}^t_{e} \in \mathbb{R}^{d}$ for interaction $e$ with $d$ as the dimension. We validate the effectiveness of the learned representations via dynamic link prediction. Specifically, the task is to decide whether node $u$ and $v$ will interact at time $t$, such that $(u,v,t)\in \mathcal{G}$.

\section{Methodology}
\label{section-3}
\begin{figure*}[!htbp]
    \centering
\includegraphics[width=2.00\columnwidth]{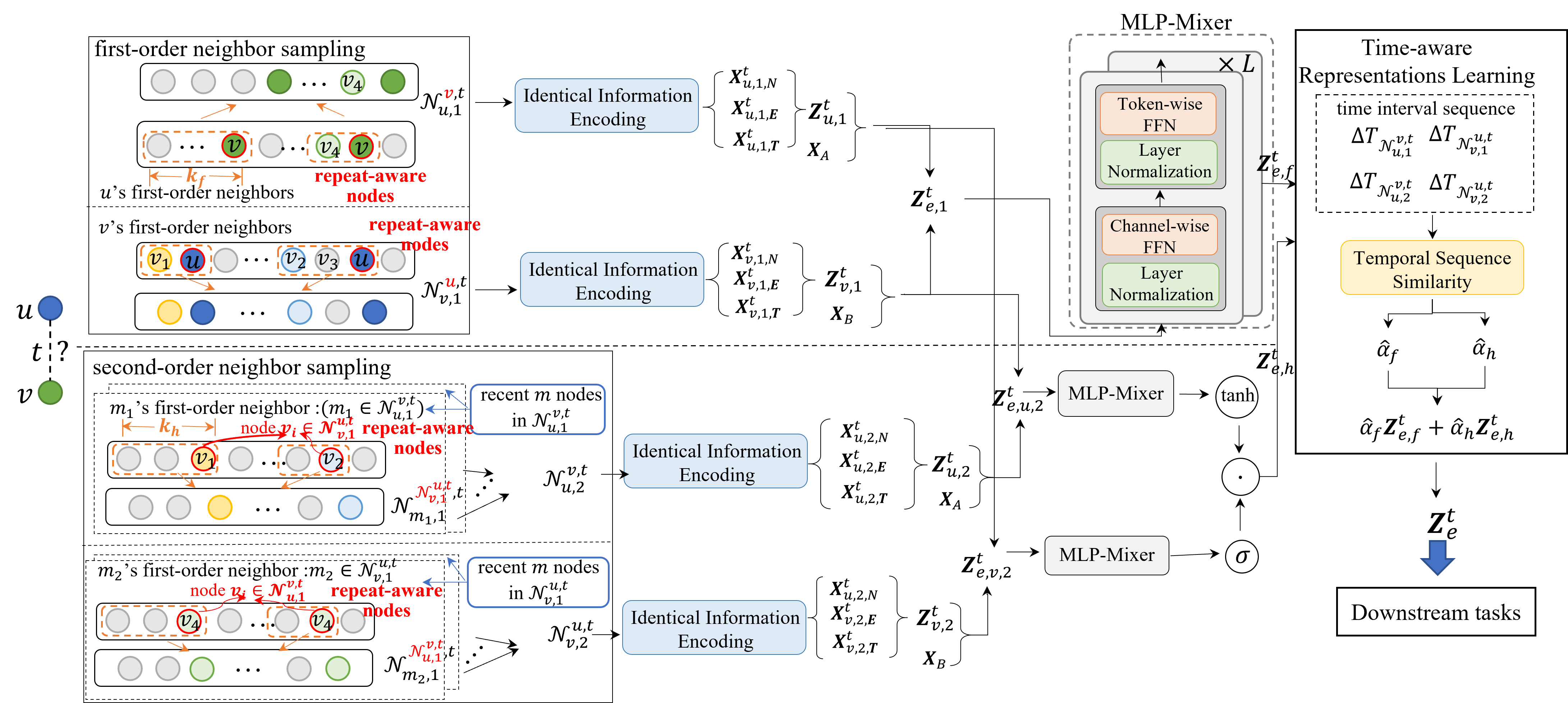}
    \caption{Framework of RepeatMixer.}
    \label{fig:framework}
\end{figure*}

\figref{fig:framework} shows the framework of our approach, which consists of three components: repeat-aware neighbor sampling strategy, RepeatMixer, and time-aware representations learning modules. Given an interaction $(u,v,t)$ that we aim to predict, we first sample the first and high-order neighbor sequence for each node, which captures the pair-wise temporal patterns among historical sequences. The second part introduces an MLP-like encoder to learn the temporal patterns of interactions based on first and higher-order neighbor sequences. Ultimately, we apply a time-aware representation learning module to generate the interaction representations from different orders at timestamp $t$ according to the significance of their time interaction sequences. Lastly, the generated representation would be used for downstream tasks in dynamic graph analysis.

\subsection{Repeat-aware Neighbor Sampling Strategy}
Selecting appropriate neighbors holds paramount significance across diverse applications, such as graph learning\cite{DBLP:conf/kdd/YoonGSNHY21}, multi-hop question answering \cite{DBLP:conf/naacl/HuangY21}, and recommendation systems \cite{DBLP:conf/recsys/DallmannZH21}. In dynamic graph learning, conventional approaches often focus on sampling neighbors exclusively from the nodes themselves, disregarding the significance of the interacted nodes. This oversight prevents models from learning repeat patterns in the neighbor sampling process. To address this limitation, we propose a novel sampling strategy explicitly considering repeat behavior between two nodes. This refined approach aims to retrieve more pertinent mutual temporal patterns, presenting a solution to enhance the effectiveness of dynamic graph learning.

\textbf{First-order Repeat-aware Neighbor Sampling Process.} Given an interaction $(u,v,t)$, we first define repeat-aware nodes of the source node as the historically interacted destination node. Then we employ a slice window strategy to select $u$'s recent $W$ neighbors that interacted before the appearance of repeat-aware nodes as our neighbor sequences for the source node $u$. This is formally expressed by $\mathcal{N}^{v,t}_{u,1}=[b|(u,b,t_\theta)\in\mathcal{G}\wedge (u,v,t_k)\in\mathcal{G} \wedge t_k \leq t \wedge 0 < P_{v,t_k}^{u} - P_{b, t_\theta}^{u} < W]$, where $P_{v,t_k}^{u}$ and $P_{b, t_\theta}^{u}$ is the appeared position of neighbor $v$ and $b$ in $u$'s historical neighbors.
Similar to the source node, we treat the repeat-aware nodes of the destination node as the source node that have interacted in the past and obtain the neighbor sequence of $v$ as $\mathcal{N}^{u,t}_{v,1}$. However, it is noteworthy that some interactions may not be present in the historical data. Hence, we acquire their most recent neighbors to form neighbor sequences.


\textbf{High-order Repeat-aware Neighbor Sampling Process.} In the higher-order sampling process, we also explore the temporal patterns and repeat behavior for interactions. To achieve this, we initiate by retrieving the $s-1$-th level neighbor sequences of nodes $u$ and $v$, denoted as $\mathcal{N}^{v,t}_{u,s-1}$ and $\mathcal{N}^{u,t}_{v,s-1}$. For node $u$, we first define the $s$-th level repeat-aware nodes as the destination node $v$'s $s-1$-th level neighbors. Hence, for the node $m \in \mathcal{N}^{v,t}_{u,s-1}$, we search the repeat-aware nodes in its historical neighbors and choose the recent $W$ neighbors that interacted before the appearance of repeat-aware nodes as $s$-th neighbor sub-sequences. We then aggregate all the neighbor sub-sequences of nodes in $u$'s $s-1$-th level as $u$'s neighbor sequences in the $s$-th level. This is formally denoted as $\mathcal{N}^{v,t}_{u,s}=[b|(m,b,t_{\gamma})\in\mathcal{G} \wedge (m,j,t_k)\in\mathcal{G} \wedge j\in \mathcal{N}^{u,t}_{v,s-1} \wedge m\in \mathcal{N}^{v,t}_{u,s-1} \wedge t_k < t \wedge 0 < P_{j,t_k}^{m} - P_{b, t_\gamma}^{m} < W]$, where $P_{j,t_k}^{m}$ and $P_{b,t_\gamma}^{m}$ denotes the appeared position of neighbor $j$ and $b$ in $m$'s historical sequences. The sampling process for generating $v$'s $s$-th neighbor sequences is similar to $u$, which treats the nodes in $\mathcal{N}^{v,t}_{u,s-1}$ as its repeat-aware nodes in the $s$-th level to obtain $\mathcal{N}^{u,t}_{v,s}$ in the $s$-th level. Considering the time complexity, we select the most recently $M$ interacted nodes in $v$'s $s-1$-th level as the repeat-aware nodes.



\subsection{RepeatMixer}
In this section, we aim to generate the temporal representations for the current interaction $e=(u,v,t)$ from the first and higher-order neighbor sequences. Specifically, we first sample the first and higher-order neighbor sequences of node $u$ and $v$ with our repeat-aware neighbor sampling strategy. Then we capture the correlated temporal information between two nodes via an MLP-based encoder based on the sampling neighbor sequences. To capture various temporal patterns of neighbors at different levels, we obtain the temporal representations from first and higher-order perspectives. Finally, we aggregate the representations from all neighbors in the sequences to obtain first and higher-order temporal representations, denoted as $\bm{Z}^t_{e,f}$ and $\bm{Z}^t_{e,h}$.



\textbf{Identical Information Encoding.} 
As introduced in \secref{section-2}, a dynamic graph is defined as a sequence of timestamped interactions associated with node features and edge features. Hence, we first embed the identical information for nodes in first and higher-order neighbor sequences with node feature, edge feature, and time interval information. For example, for node $i \in \mathcal{N}^{v,t}_{u,1}$, node and edge features are obtained from the dynamic graph $\mathcal{G}$, as $\bm{x}_i\in\mathbf{R}^{d_N}$ and $\bm{x}_{u,i}^{t_k}\in\mathbf{R}^{d_E}$. To embed the time interval information, we follow \cite{DBLP:conf/iclr/CongZKYWZTM23} and apply $\cos(\cdot)$ functions to map the time interval $\Delta{t^{\prime}}=t-t_{k}$ into a continuous vector, where $t_k$ is the timestamp of the interaction between $u$ and $i$. By modeling time interval information, we could learn the evolving patterns of nodes. The equation is as follows,
\begin{equation}
\label{equ:temporal_representation}
\begin{split}
   \bm{x}_{i,t}^T &=\sqrt{\frac{1}{d_T}}\bigg[\cos\left(w_1 \Delta{t^{\prime}}\right),
   \cos\left(w_2 \Delta{t^{\prime}}\right), \ldots,
   \cos\left(w_{d_T} \Delta{t^{\prime}} \right) \bigg],
\end{split}
\end{equation}
where $w=[w_1,\cdots,w_{d_T}]$ are fixed features $\bm{w}=\{ \alpha^{-(i-1)/\beta}\}$ and  $\bm{x}_{i,t}^T$ has a dimension of $d_T$. In this work, we use $d_T=100$ and $\alpha=\beta=\sqrt{d_T}$, following the approach outlined in \cite{DBLP:conf/iclr/CongZKYWZTM23}. 

In this work, we encode the nodes in first and second-order neighbor sequences of $u$ and $v$. Given the greater relevance of recent behaviors in capturing temporal information, we select the most recent $K$ neighbors from each sequence to learn temporal information in the subsequent sections. To ensure a consistent length of $K$ for each node's neighbor sequence, we apply zero-padding to neighbor sequences that have fewer than $K$ neighbors. As a result, the embeddings of $\mathcal{N}^{v,t}_{u,1}$ and $\mathcal{N}^{v,t}_{u,2}$ are denoted as $\bm{Z}^t_{u, 1}=\bm{X}^t_{u, 1, N} || \bm{X}^t_{u, 1, E} || \bm{X}^t_{u, 1, T} \in \mathbb{R}^{K \times d}$ and $\bm{Z}^t_{u, 2}=\bm{X}^t_{u, 2, N} || \bm{X}^t_{u, 2, E} || \bm{X}^t_{u, 2, T} \in \mathbb{R}^{K \times d}$. Similarly, the embeddings of $\mathcal{N}^{u,t}_{v,1}$ and $\mathcal{N}^{u,t}_{v,2}$ are $\bm{Z}^t_{v, 1}=\bm{X}^t_{v, 1, N} || \bm{X}^t_{v, 1, E} || \bm{X}^t_{v, 1, T} \in \mathbb{R}^{K \times d}$ and $\bm{Z}^t_{v, 2}=\bm{X}^t_{v, 2, N} || \bm{X}^t_{v, 2, E} || \bm{X}^t_{v, 2, T} \in \mathbb{R}^{K \times d}$ In short, we define $d=d_{N}+d_{E}+d_{T}$ as the dimension of initialized embeddings for $u$ and $v$ throughout the rest of our work.



\subsubsection{Representation Learning for First-order Temporal Information.}
Given an interaction $e=(u,v,t)$, we first obtain the embeddings of their first-order neighbor sequences with node features, edge interaction features, and time interval information, represented as $\bm{Z}^t_{u, 1}$ and $\bm{Z}^{t}_{v, 1}$ respectively. Then we capture the long-term dependencies and correlated structure information between two sequences via an MLP-like encoder. Lastly, we generate the first-order temporal representations $\bm{Z}_{e, f}^{t}$ for the interaction $e$ by aggregating the information from all neighbors in the sequence.

\textbf{Temporal Information Fusion.} To capture long-term temporal dependencies for nodes and correlated patterns in interactions, we merge the sequences from $u$ and $v$ and utilize an MLP-based architecture \cite{DBLP:conf/nips/TolstikhinHKBZU21} to learn the temporal interaction information. Furthermore, to facilitate the model in discerning information from node $u$ or $v$, we introduce trainable segment embeddings $\bm{x}_A,\bm{x}_B \in \mathbb{R}^{d_S}$ for each node in $\mathcal{N}^{v}_{u,1}$ and $\mathcal{N}^{u}_{v,1}$, represented as $\bm{X}_A$ and $\bm{X}_B$ for sequences $\mathcal{N}^{v}_{u,1}$ and $\mathcal{N}^{u}_{v,1}$. Subsequently, we concatenate the sequential encodings from $u$ and $v$ to form $\bm{Z}^{t}_{e, 1}=[\bm{Z}^t_{u, 1}||\bm{X}_A; \bm{Z}^t_{v, 1}||\bm{X}_B] \in \mathbb{R}^{2K \times (d+d_S)}$. Next, we derive the temporal embeddings $\bm{Z}^t_{e,f}$ via an MLP-based encoder, which is built by stacking $L$ two MLP blocks. Before each block, we add LayerNorm \cite{DBLP:journals/corr/BaKH16}, and after each block, we employ a residual connection \cite{DBLP:conf/cvpr/HeZRS16}. Additionally, we use GeLU \cite{DBLP:journals/corr/HendrycksG16} as the activation function between fully-connected layers. The process is as follows,

{\small
\begin{align}
    \bm{H}^0_{e} &= \bm{W}_e\bm{Z}^{t}_{e, 1}, \\
    \text{FFN}&(\bm{I},\bm{W}_1, \bm{b}_1, \bm{W}_2, \bm{b}_2) = \text{GeLU}(\bm{I}\bm{W}_1+\bm{b}_1)\bm{W}_2+\bm{b}_2, \\
\label{equ:MLP-Mixer}
    \bar{\bm{H}}^{l-1}_{e} &= \text{LayerNorm}_{\text{token}}(\bm{H}^{l-1}_{e}), \\
   \bm{O}^{l}_{e} &= \text{FFN}_{\text{token}}(\bar{\bm{H}}^{l-1}, \bm{W}_{o,1}^{l}, \bm{b}_{o,1}^{l}, \bm{W}_{o,2}^{l}, \bm{b}_{o,2}^{l}) + \bm{H}^{l-1}_{e}, \\
   \bar{\bm{O}}^{l}_{e} &= \text{LayerNorm}_{\text{channel}}(\bm{O}^{l}_{e}), \\
   \bm{H}^{l}_{e} &= \text{FFN}_{\text{channel}}(\bar{\bm{O}}^{l}_{e}, \bm{W}_{c,1}^{l}, \bm{b}_{c,1}^{l}, \bm{W}_{c,2}^{l}, \bm{b}_{c,2}^{l}) + \bm{O}^{l}_{e},
\end{align}
}

where we apply a transform operation to generate the initialized embeddings of the encoder, denoted as $\bm{H}^0_e\in \mathbb{R}^{2K\times d_m}$. $\bm{W}_{o,1}^{l} \in\mathbb{R}^{d_m \times d_k^o}, \bm{b}_{o,1}^{l} \in\mathbb{R}^{d_{k}^o}, \bm{W}_{o,2}^{l}\in\mathbb{R}^{d_k^o \times d_m}, \bm{b}_{o,2}^{l} \in\mathbb{R}^{d_m}, \bm{W}_{c,1}^{l} \in\mathbb{R}^{d_m \times d_k^c}$, \\$\bm{b}_{c,1}^{l} \in \mathbb{R}^{d_{k}^c}, \bm{W}_{c,2}^{l}\in\mathbb{R}^{d_k^c \times d_m}$ and $\bm{b}_{c,2}^{l} \in\mathbb{R}^{d_m}$ are trainable parameters at the $l$-th layer in the encoder. We set $d_k^o = \theta_{o}d_m$ and $d_k^c=\theta_{c}d_m$ as the dimension of hidden size. The output of the $L$-th layer is denoted by $\bm{H}^t_{e, 1}$, and we average the neighbors' representations in the concatenated sequence as local temporal representations $\bm{Z}^t_{e, f}\in \mathbb{R}^{d_m}$, which is calculated by,

\begin{equation}
    \bm{Z}^t_{e, f} = \frac{1}{K}\sum^K_{i=1} \bm{H}^t_{e, 1}[i, :].   
\end{equation}

\subsubsection{Representation Learning for High-order Temporal Information.}
In the realm of static graph learning, numerous studies \cite{DBLP:conf/icml/Abu-El-HaijaPKA19, DBLP:conf/nips/FrascaBBM22, DBLP:conf/kdd/BestaGMBKGKAGDH22,DBLP:conf/kdd/0003XL023,DBLP:conf/kdd/WuXZJSSZY22}, leverage higher-order structures to capture intricate topology information, such as triangles \cite{DBLP:conf/nips/FrascaBBM22}, motifs \cite{DBLP:conf/kdd/BestaGMBKGKAGDH22}, and communities \cite{DBLP:conf/kdd/0003XL023,DBLP:conf/kdd/WuXZJSSZY22}. In our work, we aim to capture the repeat behavior patterns in high-order neighbor sequences for learning the structure information. To strike a balance between efficiency and accuracy, we capture the second-order neighbors' temporal information in our work.

\textbf{Temporal Information Fusion.} Our definition of repeat behaviors for node $u$ at the second-order level means that they are associated with nodes of 
$v$'s first-order neighbors. Therefore, we proceed to learn the correlation between $v$'s first-order neighbors and $u$'s second-order neighbors as $u$'s higher-order temporal information. Similar to the encoding for nodes in first-order neighbors, we incorporate segment embeddings for the nodes in the neighbor sequences of $u$ and $v$. Hence, we obtain the embeddings of $u$'s higher-order temporal information as $\bm{Z}^t_{u, 2}$ and $\bm{Z}^t_{v, 1}$ as $\bm{Z}^{t}_{e, u, 2}=[\bm{Z}^t_{u,2} || \bm{X}_{A};\bm{Z}^t_{v, 1} || \bm{X}_{B}] \in \mathbb{R}^{2K\times (d+d_S)}$. Then we capture the long-term sequential and higher-order topology information in the sequences with the MLP-based encoder by stacking with $L$ MLP blocks from \equref{equ:MLP-Mixer}. Hence, we get the higher-order temporal encoding for $u$, denoted as $\bm{H}^t_{e, u, 2}\in \mathbb{R}^{2K\times d_m}$. Following the same process, we obtain the higher-order temporal encoding for $v$, denoted as $\bm{H}^t_{e, v, 2}\in \mathbb{R}^{2K\times d_m}$.

\textbf{High-order Representation Embedding.} To aggregate higher-order temporal information from two nodes, we start by gathering temporal information from neighbors in the sequence using a mean function. We then employ a gated convolution aggregation mechanism to capture information from across the two sequences, as illustrated below,
\begin{align}
\hat{\bm{H}}^t_{e, u, 2} &= \frac{1}{N}\sum_{i=1}^{K}(\bm{H}^t_{e, u, 2}[i, :]), \\
    \hat{\bm{H}}^t_{e, v, 2} &= \frac{1}{N}\sum_{i=1}^{K}(\bm{H}^t_{e, v, 2}[i, :]), \\
    \bm{Z}^t_{e,h} &= \sigma (\hat{\bm{H}}^t_{e, u,2}) \cdot \text{tanh} (\hat{\bm{H}}^t_{e, v,2}),
\end{align}
where $\hat{\bm{H}}^t_{e, u, 2}\in \mathbb{R}^{d_m}$ and $\hat{\bm{H}}^t_{e, v, 2}\in \mathbb{R}^{d_m}$ represents the aggregated higher-order representations for $u$ and $v$, and $\bm{Z}^t_{e, h}\in \mathbb{R}^{d_m}$ is the higher-order temporal representations for interaction $e=(u,v,t)$. Besides, we could also capture the information in higher-order neighbor sequences. The temporal embeddings at the higher-order level are similar to the process for second-order representations.

\subsection{Time-aware Representations Learning}
Based on the analysis in \secref{section-1}, it indicates that time interval sequences play a crucial role in capturing the evolving patterns of nodes. These sequences provide vital information regarding the frequency of interactions and behavioral patterns exhibited by the nodes. Hence, we obtain the time-aware representations of interaction $e=(u,v,t)$ by averaging the first-order representations $\bm{Z}^t_{e,f}$ and higher-order representations $\bm{Z}^t_{e,h}$ adaptively based on the significance in their time interaction sequences.

\textbf{Temporal Sequence Similarity.} To aggregate the temporal representations from both first-order and higher-order neighbors, we initially calculate the importance score between the two using a pearson correlation coefficient (PCC) similarity function. This score serves as an indicator of the similarity between the respective sequences. Subsequently, we normalize these scores and utilize them as weights to aggregate the first-order and higher-order representations together.
\begin{align}
    Cov({T}_1, {T}_2) &=  \frac{\sum_{i}^{N}(T_1[i]-\bar{T}_1)(T_2[i]-\bar{T}_2)}{N-1}, \\ 
    g({T_1,T_2}) &= \frac{Cov({T}_1, {T}_2)}{\sigma_{T_1} \sigma_{T_2}},
\end{align}
where $T_1,T_2\in \mathbb{R}^N$ are two time series with $N$ records, and $\bar{T}_1, \bar{T}_2$ are the average values of $T_1$ and $T_2$, $\sigma_{T_1}$ and $\sigma_{T_2}$ are their standard deviation. In our work, we calculate the temporal similarity of the time interval series associated with the sequential information on first-order neighbors and higher-order neighbors, denoted as $\alpha_f$ and $\alpha_h$. The process is as follows,
\begin{align}
    \alpha_f &= g(\Delta T_{\mathcal{N}^{v, t}_{u, 1}}, \Delta T_{\mathcal{N}^{u, t}_{v, 1}}), \\
    \alpha_h &= g(\Delta T_{\mathcal{N}^{v, t}_{u, 1}}||\Delta T_{\mathcal{N}^{u, t}_{v, 2}}, \Delta T_{\mathcal{N}^{u, t}_{v, 1}} || \Delta T_{\mathcal{N}^{v, t}_{u, 2}}).
\end{align}

Since we capture the second-order repeat patterns between the first- and second-order neighbors in different nodes, hence, we calculate the higher-order similarity between $\Delta T_{\mathcal{N}^{v, t}_{u, 1}}||\Delta T_{\mathcal{N}^{u, t}_{v, 2}}$ and $\Delta T_{\mathcal{N}^{u, t}_{v, 1}} || \Delta T_{\mathcal{N}^{v, t}_{u, 2}}$ to keep in consistency.

\textbf{Adaptive Time-aware Aggregation.} We then normalize the weights of $\alpha_f$ and $\alpha_h$ as weights to aggregate the first-order representations and higher-order representations for the interaction. 
\begin{align}
        \hat{\alpha}_f &= \frac{e^{\alpha_f}}{e^{\alpha_f} + e^{\alpha_h}}, \\
    \hat{\alpha}_h &= \frac{e^{\alpha_h}}{e^{\alpha_f} + e^{\alpha_h}}, \\
    \bm{Z}^{t}_{e} &=  \hat{\alpha}_f \bm{Z}^{t}_{e,f} + \hat{\alpha}_h \bm{Z}^{t}_{e,h}.
\end{align}

It is worth noticing that we could also aggregate temporal representations from higher orders beyond first and second-order. The process of calculating temporal sequence similarity among higher-order neighbors is similar to the second-order temporal representations based on the combination of adjacent time interval sequences from two nodes. Lastly, we could utilize the generated temporal representations $\bm{Z}^{t}_{e}$ for downstream tasks. 


\section{Experiments}
\label{section-4}
In this section, extensive experiments are conducted. We report the results of various recent models. We also demonstrate the superiority of RepeatMixer over existing methods and give an in-depth analysis of learning first and high-order temporal patterns via the repeat-aware sampling strategy technique and time-aware aggregation mechanism.

\begin{table*}[!htbp]
\centering
\caption{Performances for transductive dynamic link prediction with three negative sampling strategies.}
\label{tab:performance_comparison}
\resizebox{\textwidth}{!}
{
\setlength{}{}
{

\begin{tabular}{c|c|c|ccccccccccc}
\hline
Metric &NSS                   & Datasets  & JODIE        & DyRep        & TGAT         & TGN          & CAWN         & EdgeBank     & TCL          & GraphMixer   & DyGFormer    & RepeatMixer(F)     & RepeatMixer   \\ \hline
\multirow{21}{*}{AP}&\multirow{7}{*}{rnd}  & Wikipedia & 96.50 ± 0.14 & 94.86 ± 0.06 & 96.94 ± 0.06 & 98.45 ± 0.06 & 98.76 ± 0.03 & 90.37 ± 0.00 & 96.47 ± 0.16 & 97.25 ± 0.03 & \uline{ 99.03 ± 0.02 } & 99.00 ± 0.03 & \textbf{ 99.16 ± 0.02 } \\
                      & & Reddit    & 98.31 ± 0.14 & 98.22 ± 0.04 & 98.52 ± 0.02 & 98.63 ± 0.06 & 99.11 ± 0.01 & 94.86 ± 0.00 & 97.53 ± 0.02 & 97.31 ± 0.01 & \textbf{  99.22 ± 0.01 } & \uline{99.14 ± 0.01} & \textbf{ 99.22 ± 0.01 } \\
                      & & MOOC      & 80.23 ± 2.44 & 81.97 ± 0.49 & 85.84 ± 0.15 & \uline{ 89.15 ± 1.60 } & 80.15 ± 0.25 & 57.97 ± 0.00 & 82.38 ± 0.24 & 82.78 ± 0.15 & 87.52 ± 0.49 & 84.95 ± 0.26 & \textbf{ 92.76 ± 0.10 } \\
                      & & LastFM    & 70.85 ± 2.13 & 71.92 ± 2.21 & 73.42 ± 0.21 & 77.07 ± 3.97 & 86.99 ± 0.06 & 79.29 ± 0.00 & 67.27 ± 2.16 & 75.61 ± 0.24 & \uline{ 93.00 ± 0.12 } & 91.44 ± 0.05 & \textbf{ 94.14 ± 0.06 } \\
                      & & Enron     & 84.77 ± 0.30 & 82.38 ± 3.36 & 71.12 ± 0.97 & 86.53 ± 1.11 & 89.56 ± 0.09 & 83.53 ± 0.00 & 79.70 ± 0.71 & 82.25 ± 0.16 & \uline{ 92.47 ± 0.12 } & 92.07 ± 0.07 & \textbf{ 92.66 ± 0.07 } \\
                      & & UCI       & 89.43 ± 1.09 & 65.14 ± 2.30 & 79.63 ± 0.70 & 92.34 ± 1.04 & 95.18 ± 0.06 & 76.20 ± 0.00 & 89.57 ± 1.63 & 93.25 ± 0.57 & 95.79 ± 0.17 & \uline{ 96.33 ± 0.14 } & \textbf{ 96.74 ± 0.08 } \\ \cline{3-14} 
                      & & Avg. Rank & 8.00 &  9.00 & 7.50 & 4.83 & 5.00 &  9.17  & 8.83 & 7.17 & \uline{2.17} & 3.17 & \textbf{1.00}  \\ \cline{2-14}
& \multirow{7}{*}{hist} & Wikipedia & 83.01 ± 0.66 & 79.93 ± 0.56 & 87.38 ± 0.22 & 86.86 ± 0.33 & 71.21 ± 1.67 & 73.35 ± 0.00 & 89.05 ± 0.39 & \uline{ 90.90 ± 0.10 } & 82.23 ± 2.54 & \textbf{ 91.02 ± 0.59 } & 90.20 ± 1.04 \\
                      & & Reddit    & 80.03 ± 0.36 & 79.83 ± 0.31 & 79.55 ± 0.20 & 81.22 ± 0.61 & 80.82 ± 0.45 & 73.59 ± 0.00 & 77.14 ± 0.16 & 78.44 ± 0.18 & 81.57 ± 0.67 & \textbf{ 84.44 ± 0.50 } & \uline{ 83.02 ± 1.20 } \\
                      & & MOOC      & 78.94 ± 1.25 & 75.60 ± 1.12 & 82.19 ± 0.62 & 87.06 ± 1.93 & 74.05 ± 0.95 & 60.71 ± 0.00 & 77.06 ± 0.41 & 77.77 ± 0.92 & 85.85 ± 0.66 & \textbf{ 94.24 ± 0.25 } & \uline{ 92.19 ± 0.58 } \\
                      & & LastFM    & 74.35 ± 3.81 & 74.92 ± 2.46 & 71.59 ± 0.24 & 76.87 ± 4.64 & 69.86 ± 0.43 & 73.03 ± 0.00 & 59.30 ± 2.31 & 72.47 ± 0.49 & 81.57 ± 0.48 & \textbf{ 88.41 ± 0.07 } & \uline{ 86.73 ± 0.34           } \\
                      & & Enron     & 69.85 ± 2.70 & 71.19 ± 2.76 & 64.07 ± 1.05 & 73.91 ± 1.76 & 64.73 ± 0.36 & 76.53 ± 0.00 & 70.66 ± 0.39 & 77.98 ± 0.92 & 75.63 ± 0.73 & \textbf{ 88.19 ± 0.23 } & \uline{ 87.38 ± 0.18 } \\
                      & & UCI       & 75.24 ± 5.80 & 55.10 ± 3.14 & 68.27 ± 1.37 & 80.43 ± 2.12 & 65.30 ± 0.43 & 65.50 ± 0.00 & 80.25 ± 2.74 & 84.11 ± 1.35 & 82.17 ± 0.82 & \uline{ 86.41 ± 0.79 } & \textbf{ 87.23 ± 0.23 } \\ \cline{3-14} 
                      & & Avg. Rank &6.83 & 8.00 & 7.67 & 4.67 & 9.33 & 8.67 & 7.83 & 5.33 & 4.50 & \textbf{1.17} & \uline{2.00}           \\ \cline{2-14} 
& \multirow{7}{*}{ind}  & Wikipedia & 75.65 ± 0.79 & 70.21 ± 1.58 & 87.00 ± 0.16 & 85.62 ± 0.44 & 74.06 ± 2.62 & 80.63 ± 0.00 & 86.76 ± 0.72 & 88.59 ± 0.17 & 78.29 ± 5.38 & \textbf{ 88.98 ± 1.49 } & \uline{88.86 ± 0.97} \\
                      & & Reddit    & 86.98 ± 0.16 & 86.30 ± 0.26 & 89.59 ± 0.24 & 88.10 ± 0.24 & \uline{ 91.67 ± 0.24 } & 85.48 ± 0.00 & 87.45 ± 0.29 & 85.26 ± 0.11 & 91.11 ± 0.40 & \textbf{ 91.76 ± 0.56 } & 91.11 ± 0.73 \\
                      & & MOOC      & 65.23 ± 2.19 & 61.66 ± 0.95 & 75.95 ± 0.64 & 77.50 ± 2.91 & 73.51 ± 0.94 & 49.43 ± 0.00 & 74.65 ± 0.54 & 74.27 ± 0.92 & 81.24 ± 0.69 & \textbf{ 83.68 ± 0.72 } & \uline{ 83.11 ± 1.28 } \\
                      & & LastFM    & 62.67 ± 4.49 & 64.41 ± 2.70 & 71.13 ± 0.17 & 65.95 ± 5.98 & 67.48 ± 0.77 & \uline{ 75.49 ± 0.00 } & 58.21 ± 0.89 & 68.12 ± 0.33 & 73.97 ± 0.50 & \textbf{ 76.34 ± 0.30 } & 75.46 ± 0.78           \\
                      & & Enron     & 68.96 ± 0.98 & 67.79 ± 1.53 & 63.94 ± 1.36 & 70.89 ± 2.72 & 75.15 ± 0.58 & 73.89 ± 0.00 & 71.29 ± 0.32 & 75.01 ± 0.79 & 77.41 ± 0.89 & \textbf{ 84.17 ± 0.24 } & \uline{ 83.17 ± 0.50 } \\
                      & & UCI       & 65.99 ± 1.40 & 54.79 ± 1.76 & 68.67 ± 0.84 & 70.94 ± 0.71 & 64.61 ± 0.48 & 57.43 ± 0.00 & 76.01 ± 1.11 & 80.10 ± 0.51 & 72.25 ± 1.71 & \textbf{ 84.96± 0.52  } & \uline{ 84.20 ± 0.34 } \\ \cline{2-14} 
                      & & Avg. Rank &8.83 & 10.00 & 6.17 & 6.33 & 6.67 & 7.67 & 6.67 & 5.83 & 4.33 & \textbf{1.00} & \uline{2.33}           \\ \hline

\multirow{21}{*}{AUC}&\multirow{7}{*}{rnd}  & Wikipedia & 96.33 ± 0.07 & 94.37 ± 0.09 & 96.67 ± 0.07 & 98.37 ± 0.07 & 98.54 ± 0.04 & 90.78 ± 0.00 & 95.84 ± 0.18 & 96.92 ± 0.03 & \uline{ 98.91 ± 0.02 } & 98.88 ± 0.03 & \textbf{ 99.04 ± 0.01 } \\
                      & & Reddit    & 98.31 ± 0.05 & 98.17 ± 0.05 & 98.47 ± 0.02 & 98.60 ± 0.06 & 99.01 ± 0.01 & 95.37 ± 0.00 & 97.42 ± 0.02 & 97.17 ± 0.02 & \textbf{99.15 ± 0.01 } & \uline {99.06 ± 0.01} & \textbf{99.15 ± 0.01 } \\
                      & & MOOC      & 83.81 ± 2.09 & 85.03 ± 0.58 & 87.11 ± 0.19 & \uline{ 91.21 ± 1.15 } & 80.38 ± 0.26 & 60.86 ± 0.00 & 83.12 ± 0.18 & 84.01 ± 0.17 & 87.91 ± 0.58 & 85.42 ± 0.25 & \textbf{ 93.60 ± 0.39 } \\
                      & & LastFM    & 70.49 ± 1.66 & 71.16 ± 1.89 & 71.59 ± 0.18 & 78.47 ± 2.94 & 85.92 ± 0.10 & 83.77 ± 0.00 & 64.06 ± 1.16 & 73.53 ± 0.12 & \uline{ 93.05 ± 0.10 } & 91.59 ± 0.01 & \textbf{ 94.27 ± 0.02 } \\
                      & & Enron     & 87.96 ± 0.52 & 84.89 ± 3.00 & 68.89 ± 1.10 & 88.32 ± 0.99 & 90.45 ± 0.14 & 87.05 ± 0.00 & 75.74 ± 0.72 & 84.38 ± 0.21 & \uline{ 93.33 ± 0.13 } & 93.09 ± 0.14 & \textbf{ 93.47 ± 0.17 } \\
                      & & UCI       & 90.44 ± 0.49 & 68.77 ± 2.34 & 78.53 ± 0.74 & 92.03 ± 1.13 & 93.87 ± 0.08 & 77.30 ± 0.00 & 87.82 ± 1.36 & 91.81 ± 0.67 & 94.49 ± 0.26 & \uline{ 95.11 ± 0.21 } & \textbf{ 95.36 ± 0.49 } \\ \cline{3-14}  
                      & & Avg. Rank & 7.67 & 8.67 & 7.50 & 4.67 & 5.00 &  9.17 &  9.33 & 7.50 & \uline{2.17} & 3.17 & \textbf{1.00}  \\ \cline{2-14} 
& \multirow{7}{*}{hist} & Wikipedia & 80.77 ± 0.73 & 77.74 ± 0.33 & 82.87 ± 0.22 & 82.74 ± 0.32 & 67.84 ± 0.64 & 77.27 ± 0.00 & 85.76 ± 0.46 & \textbf{ 87.68 ± 0.17 } & 78.80 ± 1.95 & \uline{ 86.23 ± 0.60 } & 85.32 ± 0.70 \\
                      & & Reddit    & 80.52 ± 0.32 & 80.15 ± 0.18 & 79.33 ± 0.16 & 81.11 ± 0.19 & 80.27 ± 0.30 & 78.58 ± 0.00 & 76.49 ± 0.16 & 77.80 ± 0.12 & 80.54 ± 0.29 & \textbf{ 83.41 ± 0.21 } & \uline{ 81.95 ± 0.56 } \\
                      & & MOOC      & 82.75 ± 0.83 & 81.06 ± 0.94 & 80.81 ± 0.67 & 88.00 ± 1.80 & 71.57 ± 1.07 & 61.90 ± 0.00 & 72.09 ± 0.56 & 76.68 ± 1.40 & 87.04 ± 0.35 & \textbf{ 92.68 ± 0.33 } & \uline{ 92.09 ± 0.32 } \\
                      & & LastFM    & 75.22 ± 2.36 & 74.65 ± 1.98 & 64.27 ± 0.26 & 77.97 ± 3.04 & 67.88 ± 0.24 & 78.09 ± 0.00 & 47.24 ± 3.13 & 64.21 ± 0.73 & 78.78 ± 0.35 & \textbf{ 84.38 ± 0.21 } & \uline{ 82.35 ± 0.39 } \\
                      & & Enron     & 75.39 ± 2.37 & 74.69 ± 3.55 & 61.85 ± 1.43 & 77.09 ± 2.22 & 65.10 ± 0.34 & 79.59 ± 0.00 & 67.95 ± 0.88 & 75.27 ± 1.14 & 76.55 ± 0.52 & \textbf{ 84.78 ± 0.29 } & \uline{ 84.33 ± 0.13 } \\
                      & & UCI       & \uline{ 78.64 ± 3.50 } & 57.91 ± 3.12 & 58.89 ± 1.57 & 77.25 ± 2.68 & 57.86 ± 0.15 & 69.56 ± 0.00 & 72.25 ± 3.46 & 77.54 ± 2.02 & 76.97 ± 0.24 & 77.48 ± 0.74 & \textbf{ 78.85 ± 0.43 } \\ \cline{3-14}  
                      & & Avg. Rank &5.17 & 7.83 & 8.17 & 4.33 &  9.33 & 7.50 &  8.33 & 6.50 & 5.00 & \textbf{1.67} & \uline{2.17}             \\ \cline{2-14} 
&\multirow{7}{*}{ind}  & Wikipedia & 70.96 ± 0.78 & 67.36 ± 0.96 & 81.93 ± 0.22 & 80.97 ± 0.31 & 70.95 ± 0.95 & 81.73 ± 0.00 & 82.19 ± 0.48 & \textbf{ 84.28 ± 0.30 } & 75.09 ± 3.70 & 83.64 ± 1.53 & \uline{ 83.67 ± 0.89 } \\
                      & & Reddit    & 83.51 ± 0.15 & 82.90 ± 0.31 & \uline{ 87.13 ± 0.20 } & 84.56 ± 0.24 & \textbf{ 88.04 ± 0.29 } & 85.93 ± 0.00 & 84.67 ± 0.29 & 82.21 ± 0.13 & 86.23 ± 0.51 & 86.85 ± 0.84 & 86.45 ± 1.03 \\
                      & & MOOC      & 66.63 ± 2.30 & 63.26 ± 1.01 & 73.18 ± 0.33 & 77.44 ± 2.86 & 70.32 ± 1.43 & 48.18 ± 0.00 & 70.36 ± 0.37 & 72.45 ± 0.72 & \uline{ 80.76 ± 0.76 } & 77.03 ± 0.97 & \textbf{ 81.94 ± 0.95 } \\
                      & & LastFM    & 61.32 ± 3.49 & 62.15 ± 2.12 & 63.99 ± 0.21 & 65.46 ± 4.27 & 67.92 ± 0.44 & \textbf{ 77.37 ± 0.00 } & 46.93 ± 2.59 & 60.22 ± 0.32 & \uline{ 69.25 ± 0.36 } & 67.97 ± 0.40 & 67.73 ± 0.96 \\
                      & & Enron     & 70.92 ± 1.05 & 68.73 ± 1.34 & 60.45 ± 2.12 & 71.34 ± 2.46 & 75.17 ± 0.50 & 75.00 ± 0.00 & 67.64 ± 0.86 & 71.53 ± 0.85 & 74.07 ± 0.64 & \textbf{ 80.53 ± 0.15 } & \uline{ 80.05 ± 0.53 } \\
                      & & UCI       & 64.14 ± 1.26 & 54.25 ± 2.01 & 60.80 ± 1.01 & 64.11 ± 1.04 & 58.06 ± 0.26 & 58.03 ± 0.00 & 70.05 ± 1.86 & 74.59 ± 0.74 & 65.96 ± 1.18 & \uline{ 76.12 ± 0.99 } & \textbf{ 76.56 ± 0.30 } \\ \cline{3-14} 
                      & & Avg. Rank & 8.33 &  9.83  & 6.33 & 6.33 & 5.83 & 6.33 & 7.17 & 6.17 & 4.50 & \uline{2.67} & \textbf{2.50}              \\ \hline
\end{tabular}
}}
\end{table*}

\subsection{Experimental Settings}
\textbf{Datasets.} We leverage six publicly available real-world datasets, namely Wikipedia, Reddit, MOOC, LastFM, Enron, and UCI, collected by \cite{DBLP:conf/nips/PoursafaeiHPR22}(refer to Appendix \ref{appendix:datasets&baselines} for detailed descriptions). The dataset statistics are presented in \tabref{tab:info datasets}. Specifically, we provide the ratios of repeat behaviors, namely as "Ratio of Repeat Behaviors". Based on the statistics, it is observed that more than half of the interactions have occurred multiple times. Particularly noteworthy is that in the "Reddit" and "Enron" datasets, over 90 percent of interactions have been observed during the testing phase.

\textbf{Baselines.} In our experimentation, we compare our model against nine well-established continuous-time dynamic graph learning baselines. These baselines span various techniques, including memory networks (i.e., JODIE \cite{DBLP:conf/kdd/KumarZL19}, DyRep \cite{DBLP:conf/iclr/TrivediFBZ19}, and TGN \cite{DBLP:journals/corr/abs-2006-10637}), graph convolutions (i.e., TGAT \cite{DBLP:conf/iclr/XuRKKA20}), random walks (i.e., CAWN\cite{DBLP:conf/iclr/WangCLL021}), statistics methods (i.e., EdgeBank \cite{DBLP:conf/nips/PoursafaeiHPR22}), MLP-based models (GraphMixer \cite{DBLP:conf/iclr/CongZKYWZTM23}), and sequential models (i.e., TCL\cite{DBLP:journals/corr/abs-2105-07944}, and DyGFormer\cite{DBLP:conf/nips/0004S0L23}). The descriptions of baselines are shown in Appendix \ref{appendix:datasets&baselines}.

\textbf{Evaluation Tasks and Metrics.} Our evaluation centers on the dynamic prediction task, aligning with established methodologies in prior works \cite{DBLP:conf/iclr/XuRKKA20,DBLP:journals/corr/abs-2006-10637,DBLP:conf/iclr/WangCLL021,DBLP:conf/nips/PoursafaeiHPR22}. This task is characterized by two settings: 1) a transductive setting, where the objective is to predict future links between nodes observed during training, and 2) an inductive setting, which aims to predict future links involving previously unseen nodes. To accomplish this, we employ a multi-layer perceptron, concatenating either node representations from baselines or edge representations from our model to predict link probabilities. We choose Average Precision (AP) and Area Under the Receiver Operating Characteristic Curve (AUC-ROC) as evaluation metrics. Consistent with \cite{DBLP:conf/nips/PoursafaeiHPR22}, we evaluate our work with three negative sampling strategies: random (rnd), historical (hist), and inductive (ind). The latter two strategies are particularly challenging due to their inherent complexities, as expounded upon in \cite{DBLP:conf/nips/PoursafaeiHPR22}. Dataset splits adhere to a chronological distribution, with a 70\%, 15\%, and 15\% ratio assigned to training, validation, and testing.


\textbf{Implementation Details.} 
To ensure consistent performance comparisons, we adopt the settings and performance metrics of the baseline models as outlined in \cite{DBLP:conf/nips/0004S0L23}.  The Adam optimizer is employed, and training spans 100 epochs, with a patience of 20 during early stopping. The model achieving the best performance on the validation set is selected for testing. Across all datasets, the learning rate and batch size are set to 0.0001 and 200, respectively. Specifically, in our sampling strategy, we select the recent 10 repeat-aware nodes in first and second-order neighbor sequences for searching and the length of slide window $W$ is set as 5. The hyper-parameter of $\theta_o$ and $\theta_c$ is set as $0.4$ and $4.0$. In all models, the dimensions of node features and edge features are set to 172. The time encoding dimensions are consistent at 100 across all models. The remaining settings of the models remain unchanged as described in their respective papers. The experiments are executed on an Ubuntu machine featuring an Intel(R) Xeon(R) Gold 6130 CPU @ 2.10GHz with 16 physical cores. The GPU device utilized is the NVIDIA Tesla T4 with 15 GB memory. The number of neighbors sampled in the neighbor sequences of different models is shown in Appendix \ref{ref:settings}. Our code is available at \url{https://github.com/Hope-Rita/RepeatMixer}.

\subsection{Performance Comparison and Discussions}
Due to space limitations, we present the performance of different methods on the AP and AUC metrics for transductive dynamic link prediction using three negative sampling strategies in Table \ref{tab:performance_comparison}. Notably, the results are multiplied by 100 for improved readability. The best and second-best results are highlighted in bold and underlined fonts. Additional results on AP for both transductive and inductive link prediction tasks can be found in \tabref{tab:ap_ind_performance}. It should be noted that EdgeBank can only be evaluated for transductive dynamic link prediction, hence its results under the inductive setting are not presented. In our work, we perform RepeatMixer(F) and RepeatMixer, which utilize just first-order neighbors and incorporate second-order temporal information for generating final temporal representations for link prediction. From Table \ref{tab:performance_comparison}, and Appendix \ref{ref:total performance}, we make the following key observations:

(i) Compared to existing works that sample recent node-wise neighbors from individual nodes as neighbor sequences, such as JODIE, TGN, and GraphMixer, RepeatMixer(F) samples the first-order neighbors considering the temporal patterns of interactions, which contains correlated pair-wise temporal behavior between two nodes for temporal representation learning.

(ii) Compared to baselines that model the correlation between nodes such as DyRep, TCL, and DyGFormer, we concatenate the sequence from the source node and target node and capture the correlation between two nodes via an MLP-based model. Among baselines, DyGFormer performs best since it calculates the co-occurrence between neighbor sequences for the source and destination node, which captures the interaction frequency between two nodes. In our work, we could achieve competitive performance with lower model complexity to learn the temporal patterns. Besides, it is worth noting that RepeatMixer(F) may underperform in "MOOC" due to its lower ratio of repeat behavior for interactions for learning first-order temporal dependencies than other datasets.

(iii) Compared to models capturing high-order temporal information such as TGAT and CAWN, we incorporate the second-order neighbor sequences between nodes to capture the correlated temporal relationships in high-order dynamic graphs. RepeatMixer achieves the best performance, which illustrates the importance of capturing repeat behavior patterns in high-order neighbors. 

(iv) Compared to the performance of capturing first-order and incorporated high-order temporal correlations between nodes, we find that RepeatMixer performs well in the random negative sampling strategy, which illustrates the importance of considering correlations between nodes in high-order neighbors, particularly in bipartite graphs. However, RepeatMixer(F) gets better performance with historical and inductive sampling strategies, which indicates the interactions with first-order neighbors are enough for prediction while high-order neighbors may bring some irrelevant temporal information and make performance degrade.

\subsection{Effects of Repeat-aware Neighbor Sampling Strategy.}
In this study, we introduce the repeat-aware Neighbor Sampling Strategy (i.e., repeat-aware NSS), which aims to capture the correlations between nodes. To evaluate the ability to capture temporal patterns for interactions, we conduct three experiments from different perspectives, including generalizability, effectiveness, and stability. To reduce the effect of high-order neighbors, we just employ the neighbor sequences from the first order in this section.

\begin{figure}[!htbp]
    \centering
\includegraphics[width=1.00\columnwidth]{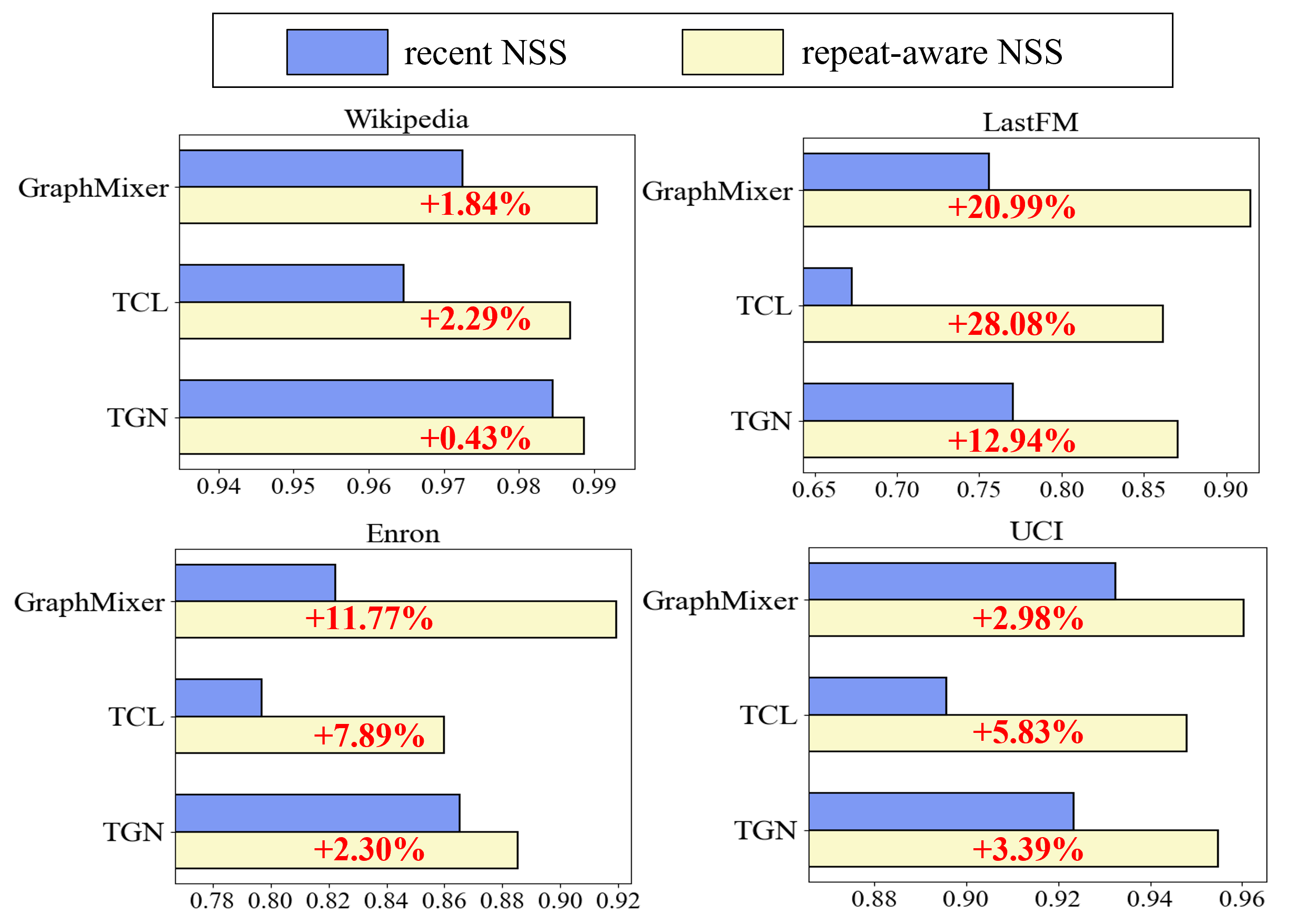}
    \caption{Performance of baselines equipped with our repeat-aware neighbor sampling strategy.}
    \label{fig:baselines std}
\end{figure}

\textbf{Generalizability of Repeat-aware Neighbor Sampling Strategy.} Our repeat-aware neighbor sampling strategy is not only specific to RepeatMixer but can also be applied to other dynamic graph learning methods that rely on sampling recent neighbors to learn temporal information, such as TGN, TCL, and GraphMixer. Hence, we replaced the original neighbor sampling strategy (i.e., recent neighbor sampling strategy) in these methods with our repeat-aware neighbor sampling strategy. The AP on the transductive link prediction is presented in \figref{fig:baselines std}. Interestingly, we observed that TGN, TCL, and GraphMixer consistently achieve superior results when using our repeat-aware neighbor sampling strategy. This finding suggests that by incorporating our sampling strategy, these models can capture a greater amount of correlated temporal information for each interaction. Among these three baselines, TGN and TCL both capture the temporal correlation between two nodes in the model, while GraphMixer learns the temporal patterns of nodes via their individual neighbor sequences. Hence, GraphMixer gets large improvement in these datasets by considering the correlated information between nodes. The consistent improvement across different methods further highlights the effectiveness and versatility of our repeat-behaviors neighbor sampling strategy.

\begin{table}[]
\caption{AP for transductive and inductive dynamic link prediction with random negative sampling strategies.}
\label{tab:sampling strategy}
\resizebox{\columnwidth}{!}
{
\setlength{}{}
{
\begin{tabular}{c|c|cccc}
\hline
                              & Sampling STR & Wikipedia    & LastFM       & Enron        & UCI          \\ \hline
\multirow{4}{*}{Trans} & uniform NSS           & 94.75 ± 0.73 & 67.15 ± 0.32 & 70.22 ± 2.08 & 89.02 ± 0.60 \\
                              & time-aware NSS        & 94.55 ± 0.67 & 66.47 ± 0.13 & 70.74 ± 1.56 & 88.87 ± 0.73 \\
                              & recent NSS            & 97.66 ± 0.24 & 82.24 ± 0.21 & 84.61 ± 1.78 & 94.51 ± 0.16 \\
                              & repeat-behaviors NSS              & \textbf{99.00 ± 0.03} & \textbf{91.44 ± 0.05} & \textbf{92.07 ± 0.07} & \textbf{96.33 ± 0.14} \\ \hline
\multirow{4}{*}{Induc}    & uniform NSS           & 93.93 ± 0.76 & 76.39 ± 0.24 & 56.54 ± 0.77 & 83.88 ± 0.73 \\
                              & time-aware NSS        & 93.77 ± 0.67 & 76.01 ± 0.16 & 56.81 ± 2.04 & 84.05 ± 1.10 \\
                              & recent NSS            & 97.20 ± 0.22 & 86.60 ± 0.33 & 77.84 ± 1.99 & 92.23 ± 0.20 \\
                              & repeat-behaviors NSS              & \textbf{98.62 ± 0.03} & \textbf{92.95 ± 0.12} & \textbf{88.16 ± 0.22} & \textbf{94.76 ± 0.08} \\ \hline
\end{tabular}
}}
\end{table}

\textbf{Comparison with Various Neighbor Sampling Strategies.} We conduct experiments to evaluate the effectiveness of our proposed NSS by comparing it with various neighbor sampling strategies, which include Recent NSS, Uniform NSS, and Time-Aware NSS. Specifically, Recent NSS selects the most recent neighbors from the historical neighbor sequences of nodes. Uniform NSS uniformly samples neighbors from the historical neighbor sequences of nodes. Time-aware NSS incorporates a parameter $\alpha$ to probabilistically select neighbors from the historical sequences, giving priority to either recent or uniform sampling. In our work, we set $\alpha$ to 0.2. The results of these experiments are summarized in \tabref{tab:sampling strategy}.

Our findings demonstrate that our proposed sampling strategy effectively captures the temporal repeat behaviors between nodes and correlations of two nodes from their neighbor sequences. Recent NSS samples recent neighbors to capture recent behaviors of nodes, but it fails to consider the temporal relationships between nodes. Uniform NSS and Time-Aware NSS, on the other hand, do not consider the continuous temporal patterns in neighbor sequences and thus yield suboptimal results.

\begin{figure}[!htbp]
    \centering
\includegraphics[width=1.00\columnwidth]{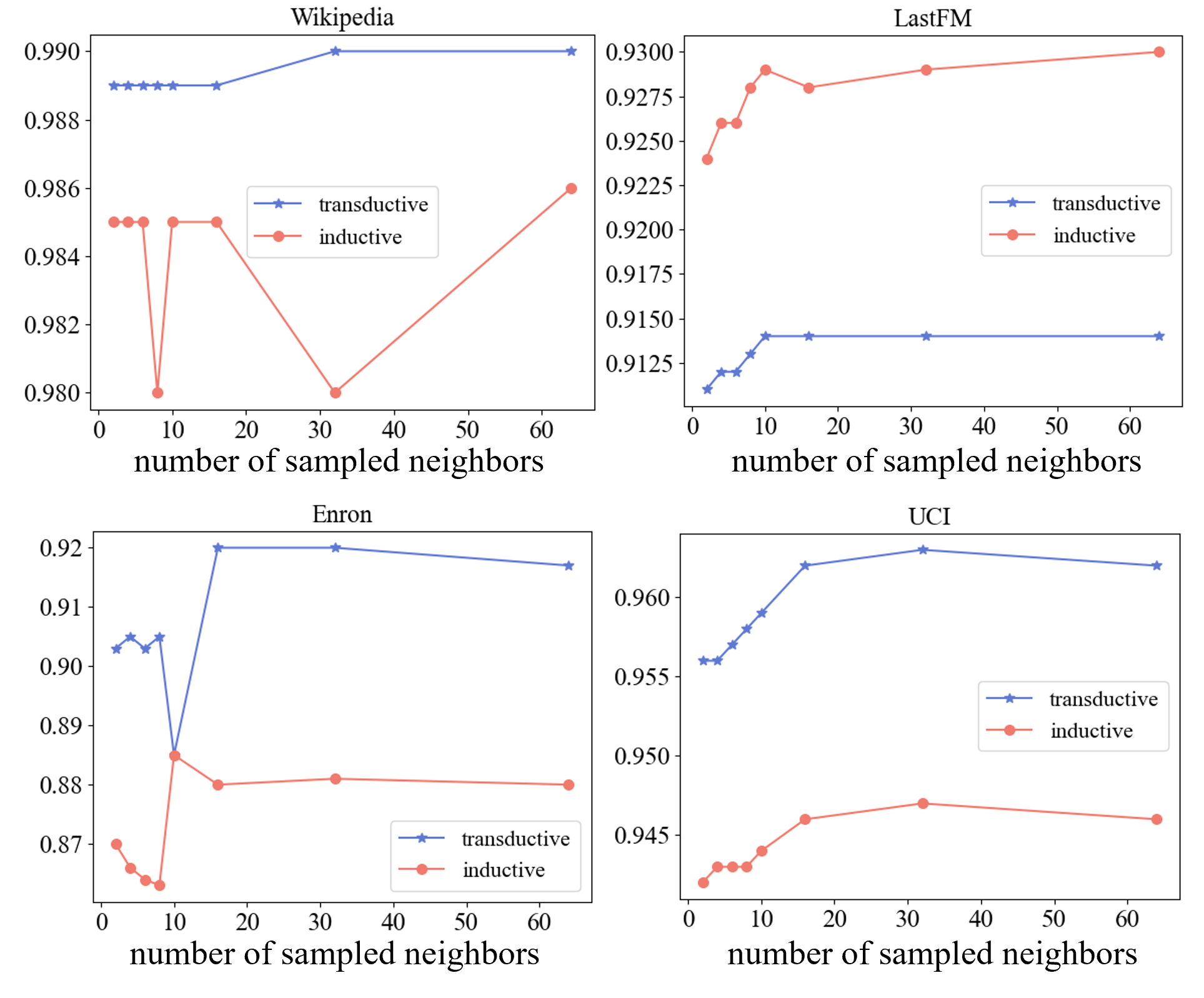}
    \caption{Effects of sampling different numbers of neighbors in neighbor sequences in RepeatMixer(F).}
    \label{fig:neighbors std}
\end{figure}

\textbf{Stability of Varied Length in Sampled Neighbor Sequence.} In this study, we utilize a repeat-aware neighbor sequence consisting of $K$ recent neighbors to extract temporal representations for interactions. To further assess the impact of different lengths of historical neighbors on capturing temporal information for nodes, we conduct experiments with varying values of $K$ in the range of $[2, 4, 6, 8, 10, 16, 32, 64]$ for transductive and inductive link prediction. The average precision is illustrated in \figref{fig:neighbors std}.

From \figref{fig:neighbors std}, it reveals that our model achieves optimal performance when the number of historical neighbors falls within a certain range. Furthermore, the performance shows relatively stable variations across datasets such as "Wikipedia," "LastFM," and "UCI" even when the number of neighbors varies significantly. This observation suggests that sampling neighbors from a pairwise perspective enables the filtering of irrelevant neighbors, facilitating the extraction of effective temporal dependencies between two nodes.



\subsection{Effects of Time-aware Aggregation Mechanism.}
In our work, we propose a time-aware aggregation mechanism that incorporates the significance of time interaction sequences between two nodes to fuse the temporal representations adaptively among different orders. To evaluate the effectiveness of adaptive aggregation according to time interaction sequences, we experiment with different kinds of fusion of first and high-order representations, including the summation and concatenation of two representations. Specifically, summation represents summing the first and high-order representations directly, i.e., $\bm{Z}^t_{e} = \bm{Z}^t_{e,f} + \bm{Z}^t_{e,h}$, namely "Summation" while concatenation is implemented with a concatenation of two representations, i.e., $\bm{Z}^t_e= [\bm{Z}^t_{e,f};\bm{Z}^t_{e,h}]$, namely "Concatenation". The performance is shown in \tabref{tab:time-aware aggregation}.

\begin{table}[]
\caption{AP for transductive and inductive dynamic link prediction with different aggregation methods.}
\label{tab:time-aware aggregation}
\resizebox{\columnwidth}{!}
{
\setlength{}{}
{
\begin{tabular}{c|ccc|ccc}
\hline
\multirow{2}{*}{Datasets} & \multicolumn{3}{c|}{Transductive}           & \multicolumn{3}{c}{Inductive}               \\ \cline{2-7} 
                          & Ours         & Summation    & Concatenation & Ours         & Summation    & Concatenation \\ \hline
Wikipedia                 & \textbf{99.16 ± 0.02} & 99.14 ± 0.02 & 99.14 ± 0.13  & 98.70 ± 0.05 & \textbf{98.75 ± 0.02} & 98.69 ± 0.04  \\
Reddit                    & 99.22 ± 0.01 & \textbf{99.23 ± 0.01} & 99.22 ± 0.02  & \textbf{98.85 ± 0.01} & 98.83 ± 0.03 & 98.84 ± 0.01  \\
MOOC                      & \textbf{92.76 ± 0.10} & 92.44 ± 0.17 & 91.75 ± 0.23  & \textbf{93.05 ± 0.12} & 91.60 ± 0.20 & 92.17 ± 0.11  \\
LastFM                    & \textbf{94.14 ± 0.06} & 93.38 ± 0.06 & 93.45 ± 0.19  & \textbf{94.98 ± 0.12} & 94.52 ± 0.17 & 94.58 ± 0.32  \\
Enron                     & \textbf{92.66 ± 0.07} & 92.25 ± 0.08 & 92.26 ± 0.20  & \textbf{87.97 ± 0.29} & 87.51 ± 0.35 & 87.26 ± 0.35  \\
UCI                       & \textbf{96.74 ± 0.08} & 96.69 ± 0.06 & 96.71 ± 0.14  & 95.04 ± 0.12 & 95.07 ± 0.06 & \textbf{95.08 ± 0.12}  \\ \hline
\end{tabular}

}}
\end{table}

\begin{figure}[!htbp]
    \centering
\includegraphics[width=1.00\columnwidth]{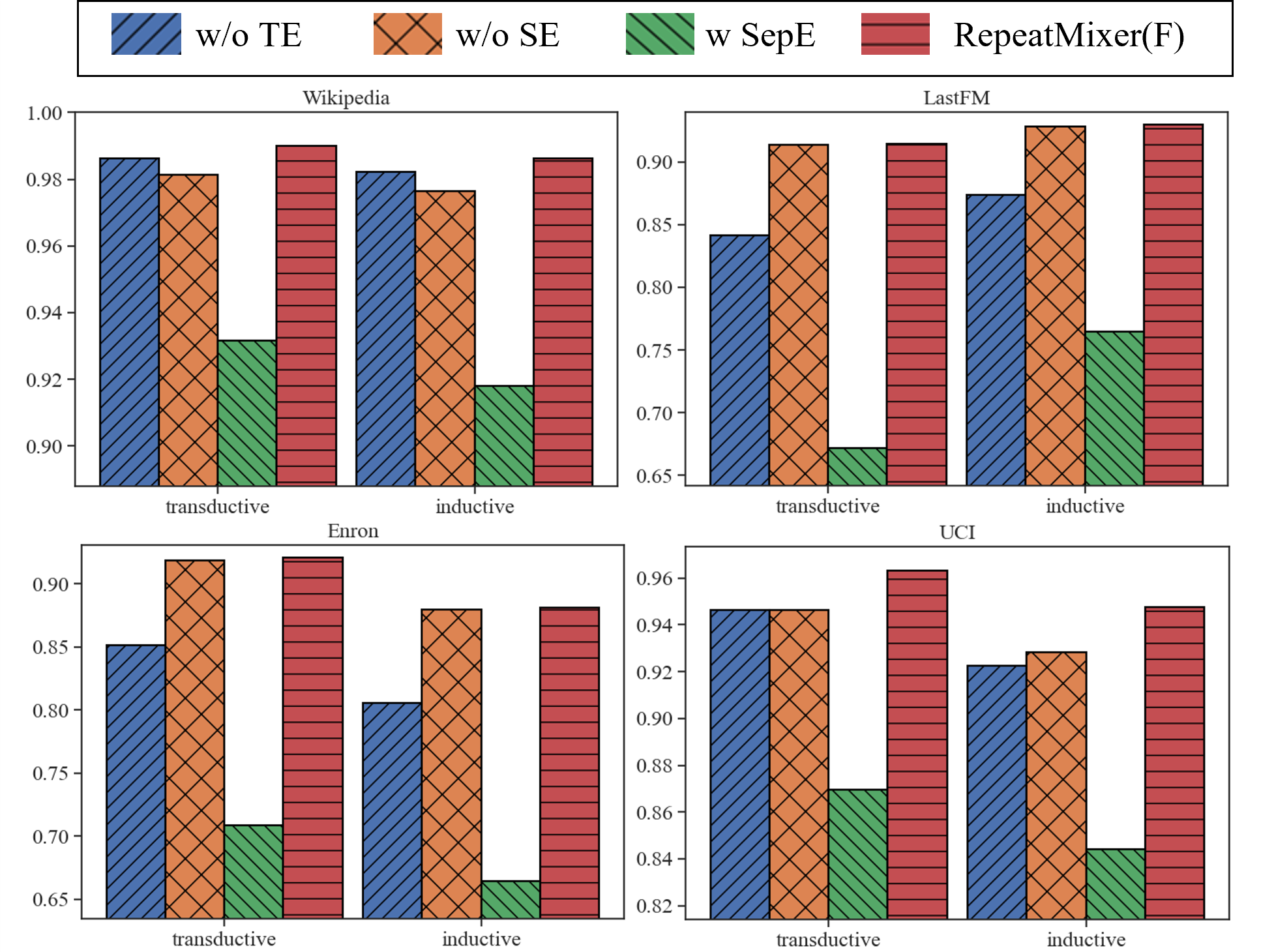}
    \caption{Effects of different components in RepeatMixer(F).}
    \label{fig:ablation std}
\end{figure}

The analysis of the results reveals that in datasets requiring high-order temporal patterns, such as "MOOC" and "LastFM," traditional aggregation methods like summation and concatenation may struggle to adaptively combine first and high-order temporal representations due to their fixed nature of aggregation. In contrast, our mechanism excels at discerning the crucial temporal patterns between the first and high-order levels. This ability allows our approach to dynamically adjust the aggregation ratios based on the importance of these temporal patterns, leading to the generation of more informative representations.

\subsection{Ablation Study}
We conduct an ablation study to further validate the effectiveness of certain designs in RepeatMixer. To assess the impact of high-order neighbor dependencies, we conduct experiments using RepeatMixer(F). We examine the impact of Time Encoding (TE) and Segment Encoding (SE) by removing these modules and denoting them as "w/o TE" and "w/o SE" respectively. Besides, we also separate the neighbors' sequence of the source node and the destination node to encode separate temporal information, denoted as "w SepE". The results are shown in \figref{fig:ablation std}.

Our findings indicate that RepeatMixer(F) achieves the best performance when all components are utilized, and the performance deteriorates when any component is removed. Particularly, the encoding of concatenation on two sequences has the most substantial impact on the model's performance, as it effectively captures the correlations between nodes. The inclusion of temporal encoding information provides valuable insights into the interaction frequency and evolving patterns of nodes, leading to improved performance. Moreover, incorporating trainable embeddings to distinguish sequences from different nodes contributes to the model's ability to learn unique temporal information.




\subsection{Experiments on Large Datasets.}
Furthermore, we conduct a thorough evaluation of our model on two large datasets, "tgbl-wiki-2" and "tgbl-review-v2", designed for dynamic link prediction in \cite{DBLP:journals/corr/abs-2307-01026}. During testing, the "tgbl-wiki-2" dataset utilizes all nodes in the graph as negative samples. Baseline performance results are obtained from \cite{DBLP:journals/corr/abs-2307-12510}. The performance is presented in \tabref{tab:large dataset}. Considering the dynamic nature of the datasets, which exhibit repeat behaviors and high temporal similarity in neighbor sequences of two nodes, our model showcases exceptional proficiency in sampling pertinent neighbors from historical sequences. This unique ability empowers us to effectively capture the temporal correlations between nodes.

\begin{table}[!htbp]
\caption{AP for dynamic link prediction on large datasets.}
\label{tab:large dataset}
\begin{tabular}{c|cc}
\hline
\textbf{Methods} & \textbf{tgbl-wiki-v2} & \textbf{tgbl-review-v2} \\ \hline
JODIE            & 63.05 ± 1.69          & \uline{41.43 ± 0.15}            \\
DyRep            & 51.91 ± 1.95          & 40.06 ± 0.59            \\
TGAT             & 59.94 ± 1.63          & 19.64 ± 0.23            \\
TGN              & 68.93 ± 0.53          & 37.48 ± 0.23            \\
CAWN             & 73.04 ± 0.60          & 19.30 ± 0.10            \\
EdgeBank         & 52.50 ± 0.00          & 2.29 ± 0.00             \\
TCL              & 78.11 ± 0.20          & 16.51 ± 1.85            \\
GraphMixer       & 59.75 ± 0.39          & 36.89 ± 1.50            \\
DyGFormer        & \textbf{79.83 ± 0.42}          & 22.39 ± 1.52            \\
RepeatMixer(F)         & \uline{79.81 ± 1.05}            & \textbf{51.37 ± 0.38}            \\ \hline
\end{tabular}
\end{table}


\section{Related Works}
\label{section-5}

\subsection{Dynamic Graph Learning}
In recent years, substantial studies \cite{DBLP:journals/jmlr/KazemiGJKSFP20, DBLP:journals/csur/BarrosMVZ23} have been proposed to explore representation learning for dynamic graphs, helping better understand evolving networks in real-life. Dynamic graph learning falls into two categories. One branch \cite{DBLP:conf/kdd/YouDL22, DBLP:conf/kdd/ZhangCFXZSC23} considers the dynamic graph as a sequence of snapshots, which treats each snapshot that contains the interactions up to a certain time. 
However, these methods have to split dynamic graphs into snapshots, 
which fails to capture fine-grained temporal information.
To tackle the problems, continuous-time approaches treat the dynamic graphs as a flow of timestamped interactions. To learn the temporal information, they either apply temporal random walks \cite{DBLP:conf/bigdataconf/NguyenLRAKK18, DBLP:conf/kdd/YuCAZCW18} to generate temporal structures for nodes or save evolving graph structures into memorable representations \cite{DBLP:conf/kdd/KumarZL19, DBLP:journals/tkde/YuLSDLL24, DBLP:journals/corr/abs-2006-10637}. 
Besides, \cite{DBLP:journals/corr/abs-2105-07944, DBLP:conf/nips/0004S0L23} employ sequential models to learn the long-term temporal dependencies.
Although dynamic graph learning has succeeded, most existing methods fail to sample repeat-aware neighbors in interaction and ignore the higher-order topology structure.

In this paper, we design a repeat-aware neighbor sampling strategy that incorporates the evolving patterns of interactions into the sampling process, which models both the first and high-order neighbor information simultaneously.


\subsection{Graph Sampling Strategy}

Graph sampling strategy \cite{DBLP:journals/ieeejas/LiuYDLYF22} is a fundamental aspect of graph analysis and processing, which reduces computational complexity while preserving the essential structural properties of the graph.
The graph sampling strategy is divided into three categories. The early works \cite{DBLP:conf/kdd/YingHCEHL18, DBLP:conf/icml/ChenZS18} collect the neighbors of all nodes in a mini-batch and then sample the entire neighborhood for the batch and proceed recursively layer by layer. 
Another branch \cite{DBLP:conf/nips/HamiltonYL17, DBLP:journals/topc/ChenSCZGC18} designs node-wise sampling that modifies the neighborhood by taking a random subset containing at most $k$ neighbors, which learns the overall distribution of nodes in the graph. 
Recent works \cite{DBLP:conf/iclr/ZengZSKP20, DBLP:conf/uai/Abu-El-HaijaDFA23} propose a graph sampling algorithm to drop boundary nodes from other partitions and ensure the connectivity among minbatch nodes. 
Our observations indicate that nodes within dynamic graphs exhibit recurrent patterns, suggesting a tendency for repeated interactions over a given period. Hence, we design a repeat-aware neighbor sampling strategy that considers the pair-wise instead of node-wise temporal information to capture the patterns.



\subsection{Repeat Behavior}
Repeat consumption \cite{DBLP:journals/tois/FangZSG20} refers to an item that repeatedly appeared in a user’s historical sequences, which is important in sequential recommendation\cite{DBLP:conf/sigir/Hu0GZ20,DBLP:conf/recsys/Reiter-HaasPSMT21, DBLP:conf/sigir/AriannezhadJLFS22}.
For example, \cite{DBLP:conf/www/AndersonKTV14} proposed that the item from $t$ timesteps ago is re-consumed with a probability proportional to a function of $t$. \cite{DBLP:conf/sigir/Hu0GZ20} designed a KNN-based model to capture repeated consumption behaviors while \cite{DBLP:conf/recsys/Reiter-HaasPSMT21} integrated a psychological theory of human cognition into re-listening music tasks. 
However, repeat behavior remains unexplored in dynamic graph learning. Hence, our work analyzes the connections between temporal interactions and repeat behavior in dynamic graphs.

\section{Conclusion}
\label{section-6}
In this paper, we proposed a dynamic graph learning method RepeatMixer with a pair-wise neighbor sampling strategy. Instead of learning the individual temporal frequency of nodes, we concentrated on the correlations between nodes in historical interactions by sampling repeat-aware neighbors, which helped us learn the evolving patterns of interactions. To obtain full temporal information, we modeled the both first and high-order neighbor sequences via an MLP-base encoder. Besides, a time-aware aggregation mechanism was introduced to adaptively fuse the temporal representations from first and high-order neighbors. Experimental results showed that our method could achieve competitive performance by learning first- and high-order node correlations via the repeat-aware neighbor sampling strategy.

\begin{acks}
This work was supported by the National Natural Science Foundation of China (62272023, 51991395, 51991391, U1811463) and the S\&T Program of Hebei (225A0802D).
\end{acks}

\bibliographystyle{ACM-Reference-Format}
\bibliography{reference}

\clearpage
\appendix
\section{Appendix}

\subsection{Datasets and Baselines}
\label{appendix:datasets&baselines}
\textbf{Datasets.} Specific statistics of datasets are shown in \tabref{tab:info datasets}. Specifically, "\#N\&L Feat" stands for the dimensions of node and link features and the ratio of repeat behaviors in training, validation, test and the whole datasets are computed in "Ratio of Repeat Behaviors". The detailed descriptions of datasets are shown as follows.
\begin{itemize}
\item \textbf{Wikipedia}: The nodes in this graph represent the users and pages, while the links indicate the connections between them. Each link is associated with a 172-dimensional Linguistic Inquiry and WordCount (LIWC) feature. Furthermore, this dataset includes dynamic labels that indicate whether users are temporarily banned from editing.

\item \textbf{Reddit}: Reddit is a bipartite graph that tracks and stores the posts made by users in various subreddits over one month. In this graph, users and subreddits are represented as nodes, while the links between them correspond to timestamped posting requests. Each link in this graph is associated with a 172-dimensional LIWC feature.

\item \textbf{MOOC}: MOOC is an online platform that operates as a bipartite interaction network, consisting of two types of nodes, namely students and course content units such as videos and problem sets. The links between these nodes represent a student's access behavior towards a specific content unit, and each link is associated with a 4-dimensional feature.

\item \textbf{LastFM}: LastFM is a bipartite system that contains data on the songs that users listened to within a one-month period. In this setup, users and songs serve as nodes, while the connections between them represent the listening habits of users.

\item \textbf{Enron}: Enron documents and archives email exchanges amongst ENRON energy corporation workers for a span of three years.

\item \textbf{UCI}: UCI is an internet-based communication network where university scholars serve as nodes and messages posted by them function as links.
\end{itemize}

\textbf{Baselines.}
\begin{itemize}
\item \textbf{JODIE}: JODIE \cite{DBLP:conf/kdd/KumarZL19} is designed for temporal bipartite networks of user-item interactions. It employs two coupled recurrent neural networks to update the states of users and items. A projection operation is introduced to learn the future representation trajectory of each user/item. 

\item \textbf{DyRep}: DyRep \cite{DBLP:conf/iclr/TrivediFBZ19} proposes a recurrent architecture to update node states upon each interaction. It also includes a temporal-attentive aggregation module to consider the temporally evolving structural information in dynamic graphs.

\item \textbf{TGN}: TGN \cite{DBLP:journals/corr/abs-2006-10637} maintains an evolving memory for each node and updates this memory when the node is observed in an interaction, which is achieved by the message function, message aggregator, and memory updater. An embedding module is leveraged to generate the temporal representations of nodes. 

\item \textbf{TGAT}: TGAT \cite{DBLP:conf/iclr/XuRKKA20} computes the node representation by aggregating features from each node’s temporal topological neighbors based on the self-attention mechanism. It is also equipped with a time encoding function for capturing temporal patterns. 

\item \textbf{CAWN}: CAWN \cite{DBLP:conf/iclr/WangCLL021} first extracts multiple causal anonymous walks for each node, which can explore the causality of network dynamics and generate relative node identities. Then, it utilizes recurrent neural networks to encode each walk and aggregates these walks to obtain the final node representation.

\item \textbf{EdgeBank}: EdgeBank \cite{DBLP:conf/nips/PoursafaeiHPR22} is a memory-based approach specifically designed for transductive dynamic link prediction. Unlike models with trainable parameters, EdgeBank operates solely on a memory unit where it stores observed interactions. Predictions are made based on whether an interaction is retained in the memory. If an interaction is stored, it is predicted as positive; otherwise, it is classified as negative.

\item \textbf{TCL}: TCL \cite{DBLP:journals/corr/abs-2105-07944} initiates the generation of each node's interaction sequence by utilizing a breadth-first search algorithm on the temporal dependency interaction sub-graph. It then introduces a graph transformer that takes into account both the graph topology and temporal information to learn node representations. Additionally, it integrates a cross-attention operation to model the interdependencies between two interaction nodes.

\item \textbf{GraphMixer}: GraphMixer \cite{DBLP:conf/iclr/CongZKYWZTM23} demonstrates that a fixed time encoding function outperforms the trainable version. It incorporates this fixed-function into a link encoder based on MLP-Mixer [54] to learn from temporal links. A node encoder using neighbor mean-pooling is implemented to summarize node features.

\item \textbf{DyGFormer}:
DyGFormer \cite{DBLP:conf/nips/0004S0L23} captures the correlations between the source node and target node via a neighbor co-occurrence encoding method based on their historical sequences and proposes a patching technique to effectively and efficiently learn longer historical neighbors by dividing the sequence into several patches.

\end{itemize}

\subsection{Settings of Models.} 
\label{ref:settings}
We show the number of sampled neighbors, causal anonymous walks, and the length of input sequences for all models at \tabref{tab:number of neighbors}.

\begin{table}[!htbp]
\caption{Configurations of the number of sampled neighbors, the number of causal anonymous walks,
 and the length of input sequences \& the patch size of different methods.}
\label{tab:number of neighbors}
\resizebox{\columnwidth}{!}
{
\setlength{}{}
{
\begin{tabular}{c|cccccccc}
\hline
Datasets & DyRep & TGAT & TGN & CAWN & TCL & GraphMixer & DyGFormer & RepeatMixer \\ \hline
Wikipedia & 10 & 20 & 10 & 32 & 20 & 30 & 32\&1 & 30 \\ 
Reddit & 10 & 20 & 10 & 32 & 20 & 10 & 64\&2 & 10 \\
MOOC & 10 & 20 & 10 & 64 & 20 & 32 & 256\&8 &  64 \\
LastFM & 10 & 20 & 10 & 128 & 20 & 10 & 512\&16 & 10 \\
Enron & 10 & 20 & 10 & 32 & 20 & 32 & 256\&8 & 16 \\
UCI & 10 & 20 & 10 & 64 & 20 & 20 & 32\&1 & 32 \\ \hline
\end{tabular}
}}
\end{table}

\subsection{Performance in transductive and inductive link prediction.}
\label{ref:total performance}
We show the performance of AP and AUC in inductive link prediction in \tabref{tab:ap_ind_performance}.

\begin{table*}[]
\centering
\caption{Information about datasets.}
\label{tab:info datasets}
\resizebox{\textwidth}{!}
{
\setlength{}{}
{

\begin{tabular}{c|ccccccccc}
\hline
Datasets  & Domains     & \#Nodes & \#Links   & \#N\&L Feat & Bipartite & Duration   & Unique Steps & Time  Granularity & Ratio of Repeat Behaviors       \\ \hline
Wikipedia & Social      & 9,227   & 157,474   & –\& 172     & True      & 1  month   & 152,757      & Unix timestamps   & 87.44\%/88.42\%/88.35\%/87.79\% \\
Reddit    & Social      & 10,984  & 672,447   & –\& 172     & True      & 1  month   & 669,065      & Unix timestamps   & 86.82\%/88.92\%/92.01\%/88.06\% \\
MOOC      & Interaction & 7,144   & 411,749   & –\& 4       & True      & 17  months & 345,600      & Unix timestamps   & 55.61\%/54.90\%/59.13\%/56.10\% \\
LastFM    & Interaction & 1,980   & 1,293,103 & –\& –       & True      & 1  month   & 1,283,614    & Unix timestamps   & 87.14\%/86.44\%/89.85\%/87.49\% \\
Enron     & Social      & 184     & 125,235   & –\& –       & False     & 3  years   & 22,632       & Unix timestamps   & 97.70\%/95.65\%/97.32\%/97.27\% \\
UCI       & Social      & 1,899   & 59,835    & –\&–        & False     & 196  days  & 58,911       & Unix timestamps   & 91.00\%/60.52\%/70.85\%/65.83\% \\ \hline
\end{tabular}
}}
\end{table*}

\begin{table*}[!htbp]
 \renewcommand{\arraystretch}{0.98}
\centering
\caption{Performance for inductive dynamic link prediction with three negative sampling strategies.}
\label{tab:ap_ind_performance}
\resizebox{\textwidth}{!}
{
\setlength{}{}
{

\begin{tabular}{c|c|c|cccccccccc}
\hline
Metric & NSS                   & Datasets  & JODIE        & DyRep        & TGAT         & TGN          & CAWN         & TCL          & GraphMixer   & DyGFormer    & RepeatMixer(F)     & RepeatMixer   \\ \hline
\multirow{21}{*}{AP} &\multirow{7}{*}{rnd}  & Wikipedia & 94.82 ± 0.20 & 92.43 ± 0.37 & 96.22 ± 0.07 & 97.83 ± 0.04 & 98.24 ± 0.03 & 96.22 ± 0.17 & 96.65 ± 0.02 & 98.59 ± 0.03 & \uline{ 98.62 ± 0.03 } & \textbf{ 98.70 ± 0.05 } \\
                     & & Reddit    & 96.50 ± 0.13 & 96.09 ± 0.11 & 97.09 ± 0.04 & 97.50 ± 0.07 & 98.62 ± 0.01 & 94.09 ± 0.07 & 95.26 ± 0.02 & \uline{ 98.84 ± 0.02 } & 98.68 ± 0.03 & \textbf{ 98.85 ± 0.01 } \\
                      & & MOOC      & 79.63 ± 1.92 & 81.07 ± 0.44 & 85.50 ± 0.19 & \uline{ 89.04 ± 1.17 } & 81.42 ± 0.24 & 80.60 ± 0.22 & 81.41 ± 0.21 & 86.96 ± 0.43 & 84.54 ± 0.31 & \textbf{ 93.05 ± 0.12 } \\
                      & & LastFM    & 81.61 ± 3.82 & 83.02 ± 1.48 & 78.63 ± 0.31 & 81.45 ± 4.29 & 89.42 ± 0.07 & 73.53 ± 1.66 & 82.11 ± 0.42 & \uline{ 94.23 ± 0.09 } & 92.95 ± 0.12 & \textbf{ 94.98 ± 0.12 } \\
                      & & Enron     & 80.72 ± 1.39 & 74.55 ± 3.95 & 67.05 ± 1.51 & 77.94 ± 1.02 & 86.35 ± 0.51 & 76.14 ± 0.79 & 75.88 ± 0.48 & \textbf{ 89.76 ± 0.34 } & \uline{ 88.16 ± 0.22 } & 87.97 ± 0.29 \\
                      & & UCI       & 79.86 ± 1.48 & 57.48 ± 1.87 & 79.54 ± 0.48 & 88.12 ± 2.05 & 92.73 ± 0.06 & 87.36 ± 2.03 & 91.19 ± 0.42 & 94.54 ± 0.12 & \uline{ 94.76 ± 0.08 } & \textbf{ 95.04 ± 0.12 } \\ \cline{3-13} 
                      & & Avg. Rank & 7.67 &   8.33 & 7.50 & 5.33 & 4.33 &  8.33 & 6.83 & \uline{2.33} & 2.83 & \textbf{1.33}              \\ \cline{2-13}
& \multirow{7}{*}{hist} & Wikipedia & 68.69 ± 0.39 & 62.18 ± 1.27 & 84.17 ± 0.22 & 81.76 ± 0.32 & 67.27 ± 1.63 & 82.20 ± 2.18 & \textbf{ 87.60 ± 0.30 } & 71.42 ± 4.43 & \uline{ 85.54 ± 1.14 } & 84.11 ± 1.88 \\
                      & & Reddit    & 62.34 ± 0.54 & 61.60 ± 0.72 & 63.47 ± 0.36 & 64.85 ± 0.85 & 63.67 ± 0.41 & 60.83 ± 0.25 & 64.50 ± 0.26 & 65.37 ± 0.60 & \textbf{ 67.45 ± 1.37 } & \uline{ 66.14 ± 1.40 } \\
                      & & MOOC      & 63.22 ± 1.55 & 62.93 ± 1.24 & 76.73 ± 0.29 & 77.07 ± 3.41 & 74.68 ± 0.68 & 74.27 ± 0.53 & 74.00 ± 0.97 & 80.82 ± 0.30 & \textbf{ 83.67 ± 0.56 } & \uline{ 83.19 ± 1.72 } \\
                      & & LastFM    & 70.39 ± 4.31 & 71.45 ± 1.76 & 76.27 ± 0.25 & 66.65 ± 6.11 & 71.33 ± 0.47 & 65.78 ± 0.65 & 76.42 ± 0.22 & 76.35 ± 0.52 & \textbf{ 81.23 ± 0.02 } & \uline{ 81.12 ± 0.30 } \\
                      & & Enron     & 65.86 ± 3.71 & 62.08 ± 2.27 & 61.40 ± 1.31 & 62.91 ± 1.16 & 60.70 ± 0.36 & 67.11 ± 0.62 & 72.37 ± 1.37 & 67.07 ± 0.62 & \textbf{ 84.49 ± 0.43 } & \uline{ 82.86 ± 0.47 } \\
                      & & UCI       & 63.11 ± 2.27 & 52.47 ± 2.06 & 70.52 ± 0.93 & 70.78 ± 0.78 & 64.54 ± 0.47 & 76.71 ± 1.00 & 81.66 ± 0.49 & 72.13 ± 1.87 & \uline{ 85.49 ± 0.20 } & \textbf{ 85.52 ± 0.16 } \\ \cline{3-13} 
                      & & Avg. Rank &  8.00  &  8.83  & 6.00 & 6.00 & 7.67 & 6.67 & 3.83 & 4.50 & \textbf{1.33} & \uline{2.17}             \\ \cline{2-13}
& \multirow{7}{*}{ind}  & Wikipedia & 68.70 ± 0.39 & 62.19 ± 1.28 & 84.17 ± 0.22 & 81.77 ± 0.32 & 67.24 ± 1.63 & 82.20 ± 2.18 & \textbf{ 87.60 ± 0.29 } & 71.42 ± 4.43 & \uline{ 85.54 ± 1.13 } & 84.10 ± 1.88 \\
                      & & Reddit    & 62.32 ± 0.54 & 61.58 ± 0.72 & 63.40 ± 0.36 & 64.84 ± 0.84 & 63.65 ± 0.41 & 60.81 ± 0.26 & 64.49 ± 0.25 & 65.35 ± 0.60 & \textbf{ 67.43 ± 1.36 } & \uline{ 66.13 ± 1.40 } \\
                      & & MOOC      & 63.22 ± 1.55 & 62.92 ± 1.24 & 76.72 ± 0.30 & 77.07 ± 3.40 & 74.69 ± 0.68 & 74.28 ± 0.53 & 73.99 ± 0.97 & 80.82 ± 0.30 & \textbf{ 83.67 ± 0.56 } & \uline{ 83.19 ± 1.72 } \\
                      & & LastFM    & 70.39 ± 4.31 & 71.45 ± 1.75 & 76.28 ± 0.25 & 69.46 ± 4.65 & 71.33 ± 0.47 & 65.78 ± 0.65 & 76.42 ± 0.22 & 76.35 ± 0.52 & \textbf{ 81.23 ± 0.02 } & \uline{ 81.12 ± 0.30 } \\
                      & & Enron     & 65.86 ± 3.71 & 62.08 ± 2.27 & 61.40 ± 1.30 & 62.90 ± 1.16 & 60.72 ± 0.36 & 67.11 ± 0.62 & 72.37 ± 1.38 & 67.07 ± 0.62 & \textbf{ 84.48 ± 0.43 } & \uline{ 82.87 ± 0.47 } \\
                      & & UCI       & 63.16 ± 2.27 & 52.47 ± 2.09 & 70.49 ± 0.93 & 70.73 ± 0.79 & 64.54 ± 0.47 & 76.65 ± 0.99 & 81.64 ± 0.49 & 72.13 ± 1.86 & \uline{ 85.48 ± 0.20 } & \textbf{ 85.53 ± 0.16 } \\ \cline{3-13} 
                      & & Avg. Rank &  8.00  &  8.83  & 6.00 & 6.00 & 7.67 & 6.67 & 3.83 & 4.50 & \textbf{1.33} & \uline{2.17}              \\ \hline

\multirow{21}{*}{AUC} & \multirow{7}{*}{rnd}  & Wikipedia & 94.33 ± 0.27 & 91.49 ± 0.45 & 95.90 ± 0.09 & 97.72 ± 0.03 & 98.03 ± 0.04 & 95.57 ± 0.20 & 96.30 ± 0.04 & \uline{ 98.48 ± 0.03 } & \uline{ 98.48 ± 0.03 } & \textbf{ 98.63 ± 0.02 } \\
                      & & Reddit    & 96.52 ± 0.13 & 96.05 ± 0.12 & 96.98 ± 0.04 & 97.39 ± 0.07 & 98.42 ± 0.02 & 93.80 ± 0.07 & 94.97 ± 0.05 & \textbf{ 98.71 ± 0.01 } & 98.53 ± 0.04 & \uline{ 98.69 ± 0.02 } \\
                      & & MOOC      & 83.16 ± 1.30 & 84.03 ± 0.49 & 86.84 ± 0.17 & \uline{ 91.24 ± 0.99 } & 81.86 ± 0.25 & 81.43 ± 0.19 & 82.77 ± 0.24 & 87.62 ± 0.51 & 85.16 ± 0.24 & \textbf{ 94.10 ± 0.07 } \\
                      & & LastFM    & 81.13 ± 3.39 & 82.24 ± 1.51 & 76.99 ± 0.29 & 82.61 ± 3.15 & 87.82 ± 0.12 & 70.84 ± 0.85 & 80.37 ± 0.18 & \uline{ 94.08 ± 0.08 } & 92.81 ± 0.02 & \textbf{ 95.10 ± 0.03 } \\
                      & & Enron     & 81.96 ± 1.34 & 76.34 ± 4.20 & 64.63 ± 1.74 & 78.83 ± 1.11 & 87.02 ± 0.50 & 72.33 ± 0.99 & 76.51 ± 0.71 & \textbf{ 90.69 ± 0.26 } & 88.85 ± 0.32 & \uline{ 89.04 ± 0.26 } \\
                      & & UCI       & 78.80 ± 0.94 & 58.08 ± 1.81 & 77.64 ± 0.38 & 86.68 ± 2.29 & 90.40 ± 0.11 & 84.49 ± 1.82 & 89.30 ± 0.57 & 92.63 ± 0.13 & \uline{ 93.08 ± 0.13 } & \textbf{ 93.54 ± 0.14 } \\ \cline{3-13} 
                      & & Avg. Rank & 7.17 &  8.00 & 7.50 & 4.83 & 4.83 &  9.00  & 7.17 & \uline{2.00} & 3.00 & \textbf{1.33}              \\ \cline{2-13}
&\multirow{7}{*}{hist} & Wikipedia & 61.86 ± 0.53 & 57.54 ± 1.09 & 78.38 ± 0.20 & 75.75 ± 0.29 & 62.04 ± 0.65 & \uline{ 79.79 ± 0.96 } & \textbf{ 82.87 ± 0.21 } & 68.33 ± 2.82 & 78.66 ± 0.82 & 77.58 ± 1.00 \\
                      & & Reddit    & 61.69 ± 0.39 & 60.45 ± 0.37 & 64.43 ± 0.27 & 64.55 ± 0.50 & 64.94 ± 0.21 & 61.43 ± 0.26 & 64.27 ± 0.13 & 64.81 ± 0.25 & \textbf{ 66.44 ± 0.66 } & \uline{ 65.15 ± 0.36 } \\
                      & & MOOC      & 64.48 ± 1.64 & 64.23 ± 1.29 & 74.08 ± 0.27 & 77.69 ± 3.55 & 71.68 ± 0.94 & 69.82 ± 0.32 & 72.53 ± 0.84 & \uline{ 80.77 ± 0.63 } & 77.42 ± 0.73 & \textbf{ 82.68 ± 1.28 } \\
                      & & LastFM    & 68.44 ± 3.26 & 68.79 ± 1.08 & 69.89 ± 0.28 & 66.99 ± 5.62 & 67.69 ± 0.24 & 55.88 ± 1.85 & 70.07 ± 0.20 & 70.73 ± 0.37 & \uline{ 72.80 ± 0.07 } & \textbf{ 73.48 ± 0.36 } \\
                      & & Enron     & 65.32 ± 3.57 & 61.50 ± 2.50 & 57.84 ± 2.18 & 62.68 ± 1.09 & 62.25 ± 0.40 & 64.06 ± 1.02 & 68.20 ± 1.62 & 65.78 ± 0.42 & \textbf{ 80.06 ± 0.32 } & \uline{ 78.84 ± 0.29 } \\
                      & & UCI       & 60.24 ± 1.94 & 51.25 ± 2.37 & 62.32 ± 1.18 & 62.69 ± 0.90 & 56.39 ± 0.10 & 70.46 ± 1.94 & 75.98 ± 0.84 & 65.55 ± 1.01 & \uline{ 76.32 ± 0.50 } & \textbf{ 76.91 ± 0.23 } \\ \cline{3-13} 
                      & & Avg. Rank & 7.67 &  9.17 & 6.17 & 6.00 & 7.17 & 6.50 & 4.00 & 4.17 & \uline{2.17} & \textbf{2.00}              \\ \cline{2-13}
&\multirow{7}{*}{ind}  & Wikipedia & 61.87 ± 0.53 & 57.54 ± 1.09 & 78.38 ± 0.20 & 75.76 ± 0.29 & 62.02 ± 0.65 & \uline{ 79.79 ± 0.96 } & \textbf{ 82.88 ± 0.21 } & 68.33 ± 2.82 & 78.66 ± 0.81 & 77.58 ± 1.00 \\
                      & & Reddit    & 61.69 ± 0.39 & 60.44 ± 0.37 & 64.39 ± 0.27 & 64.55 ± 0.50 & 64.91 ± 0.21 & 61.36 ± 0.26 & 64.27 ± 0.13 & 64.80 ± 0.25 & \textbf{ 66.42 ± 0.65 } & \uline{ 65.15 ± 0.36 } \\
                      & & MOOC      & 64.48 ± 1.64 & 64.22 ± 1.29 & 74.07 ± 0.27 & 77.68 ± 3.55 & 71.69 ± 0.94 & 69.83 ± 0.32 & 72.52 ± 0.84 & \uline{ 80.77 ± 0.63 } & 77.43 ± 0.73 & \textbf{ 82.68 ± 1.28 } \\
                      & & LastFM    & 68.44 ± 3.26 & 68.79 ± 1.08 & 69.89 ± 0.28 & 66.99 ± 5.61 & 67.68 ± 0.24 & 55.88 ± 1.85 & 70.07 ± 0.20 & 70.73 ± 0.37 & \textbf{ 7280 ± 0.07  } & \uline{ 73.48 ± 0.36 } \\
                      & & Enron     & 65.32 ± 3.57 & 61.50 ± 2.50 & 57.83 ± 2.18 & 62.68 ± 1.09 & 62.27 ± 0.40 & 64.05 ± 1.02 & 68.19 ± 1.63 & 65.79 ± 0.42 & \textbf{ 80.06 ± 0.32 } & \uline{ 78.85 ± 0.29 } \\
                      & & UCI       & 60.27 ± 1.94 & 51.26 ± 2.40 & 62.29 ± 1.17 & 62.66 ± 0.91 & 56.39 ± 0.11 & 70.42 ± 1.93 & 75.97 ± 0.85 & 65.58 ± 1.00 & \uline{ 76.31 ± 0.51 } & \textbf{ 76.92 ± 0.23 } \\ \cline{3-13} 
                      & & Avg. Rank &  7.67 &  9.17 & 6.17 & 6.00 & 7.17 & 6.50 & 4.00 & 4.17 & \textbf{2.00} & \uline{2.17}              \\ \hline
\end{tabular}
}}
\end{table*}

\subsection{Complexity Analysis.}
RepeatMixer is a sequence-based model based on a repeat-aware neighbor sampling strategy. In the neighbor sampling strategy, we first recognize the repeat-aware nodes in neighbor sequences, then select the recent neighbors before repeat-aware nodes. In the first-order sampling process, since the repeat-aware nodes are the source nodes and destination nodes, the complexity is $O(L)$ since it will compare over the whole neighbor sequence for source nodes and destination nodes. In the second-order sampling process, we select the recent $M$ one-hop neighbors of source nodes or destination nodes as the repeat-aware nodes, hence, the complexity is $(ML^2)$ since we have to search all the neighbors sequences of source nodes' first-order neighbors and destination nodes' first-order neighbors. Therefore, the whole-time complexity of our strategy both in the first-order and second-order sampling process is $O(L+ML^2)$. Since RepeatMixer is with MLPMixer encoder, the time complexity of encoding features is $O(LNd)+ O(dNL)=O(2LNd)$, where $O(dL)$ and $O(Ld)$ are time complexity for channel mixer and token mixer. Hence, the total complexity for our model is $O(L+ML^2+LNd)$.

\end{document}